\documentclass[sigconf]{acmart}

\AtBeginDocument{%
  \providecommand\BibTeX{{%
    \normalfont B\kern-0.5em{\scshape i\kern-0.25em b}\kern-0.8em\TeX}}}

\acmISBN{978-1-4503-XXXX-X/2026/06}

\usepackage{lmodern}
\usepackage[framemethod=tikz]{mdframed}
\usepackage{amsmath}
\usepackage{amsfonts} 
\usepackage{latexsym}
\usepackage{graphicx}
\usepackage{stackengine}
\usepackage{booktabs}
\usepackage[utf8]{inputenc}
\usepackage[table]{colortbl,color}
\usepackage{xcolor}
\usepackage{multirow}
\usepackage{enumitem}
\usepackage{algorithm, algorithmic}

\usepackage{pifont}

\renewcommand\footnotetextcopyrightpermission[1]{}
\begin{document}

\title{Mixture of Debaters: Learn to Debate at Architectural Level in Multi-Agent Reasoning}


\author{Dayong Liang}
\email{ft_ldy@mail.scut.edu.cn}
\affiliation{
  \institution{South China University of Technology}
  \city{Guangzhou}
  \country{China}
}

\author{Kaisong Gong}
\email{3020244013@tju.edu.cn}
\affiliation{
  \institution{Tianjin University}
  \city{Tianjin}
  \country{China}
}

\author{Yi Cai}
\email{ycai@scut.edu.cn}
\affiliation{
  \institution{South China University of Technology}
  \city{Guangzhou}
  \country{China}
}

\author{Changmeng Zheng}
\email{changmeng.zheng@polyu.edu.hk}
\authornote{Corresponding authors.}
\affiliation{
  \institution{The Hong Kong Polytechnic University}
  \city{Hong Kong}
  \country{China}
}

\author{Xiao-Yong Wei}
\email{x1wei@polyu.edu.hk}
\authornotemark[1]
\affiliation{
  \institution{The Hong Kong Polytechnic University}
  \city{Hong Kong}
  \country{China}
}


\begin{abstract}
Existing multi-agent debate frameworks suffer from two critical limitations: they rely on static architectures where agent roles and coordination patterns are fixed at design time, and they require instantiating multiple model copies, incurring substantial computational overhead. We propose \textbf{Mixture of Debaters (MoD)}, a unified framework that enables dynamic self-debate within a single model by leveraging the Mixture-of-Experts paradigm. We address three key challenges in adapting MoE for dialectical reasoning: (1) \emph{dual-routing} that decouples role allocation from process flow, dynamically determining when to debate versus when to synthesize; (2) \emph{momentum switching} that smooths token-level routing with local context, reducing expert-switch jitter; and (3) \emph{unified self-debate} that encapsulates diverse debating personas into lightweight expert modules, eliminating inter-agent communication while preserving behavioral diversity. Extensive experiments on multimodal benchmarks demonstrate that MoD outperforms both single-model baselines and conventional multi-agent systems, achieving superior accuracy with 3.7$\times$ lower latency and 87\% reduction in token consumption.
The source code can be accessed at \textcolor{magenta}{\url{https://github.com/YongLD/MoD}.}
\end{abstract}



\maketitle

\section{Introduction}
\label{sec:intro}

Large Language Models (LLMs) have demonstrated remarkable capabilities in complex reasoning tasks, ranging from mathematical problem solving to logical deduction and code generation \cite{didolkar2024metacognitive, seals2024evaluating,wei2022chain, peng2025aligning}. Despite the success of prompting strategies like Chain-of-Thought (CoT), reasoning traces generated by a single monolithic model remain prone to degeneration and hallucination, particularly when navigating intricate logical landscapes \cite{zheng2026multimodal}. To address these limitations, recent research has pivoted toward multi-agent debate frameworks, where diverse agents improve reasoning fidelity through adversarial critique and cross-examination \cite{liang2024encouraging, li2025advancing,zheng2024picture}.

\begin{figure}[t]
    \centering
    \includegraphics[width=1.0\columnwidth]{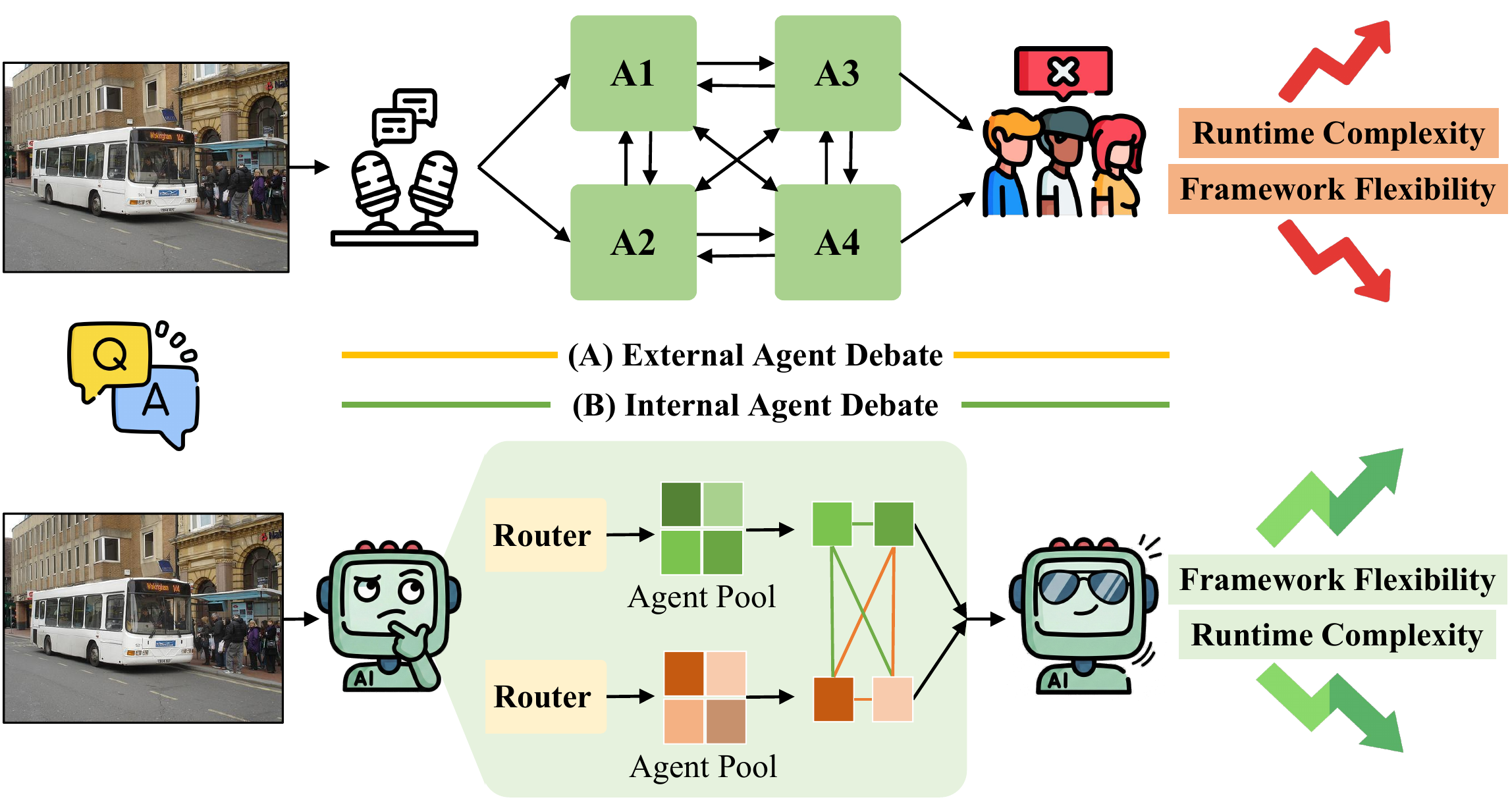}
    \Description{Internal vs. external debate architectures. (A) Conventional multi-agent debate requires instantiating independent model copies ($A_1$--$A_4$), incurring substantial runtime complexity. (B) Our Mixture-of-Debaters internalizes diverse agents into a shared expert pool with dynamic routing, improving framework flexibility while eliminating inter-agent communication overhead.}
    \caption{Internal vs. external debate architectures. (A) Conventional multi-agent debate requires instantiating independent model copies ($A_1$--$A_4$), incurring substantial runtime complexity. (B) Our Mixture-of-Debaters internalizes diverse agents into a shared expert pool with dynamic routing, improving framework flexibility while eliminating inter-agent communication overhead.}
    \label{fig:ablation_arch}
\end{figure}

However, we argue that current debate paradigms are structurally constrained. Existing systems operate with static architectures where agent roles and communication patterns are fixed at design time. For instance, a standard setup might rigidly enforce a cycle of ``propose $\rightarrow$ critique $\rightarrow$ refine.'' Real-world multimodal problems, however, are fluid and multifaceted: an advanced physics problem might initially appear as a simple kinematics question, evolve into a complex thermodynamics puzzle upon closer inspection of a diagram, and ultimately require mathematical synthesis. In such dynamic contexts, a fixed topology fails to adapt to the shifting epistemic needs of the problem.

To bridge the gap between static architectures and fluid reasoning requirements, the Mixture-of-Experts (MoE) framework offers a promising structural foundation \cite{zhou2022mixture,liu2023moelora,li2025uni}. By dynamically activating different subsets of parameters (experts) based on the input, MoE has the potential to realize robust and generalized reasoning. Nevertheless, directly applying generic MoE architectures to the dialectical process of debate is non-trivial and exposes distinct architectural gaps.

In this work, we propose \textbf{Mixture-of-Debaters} (\textbf{MoD}), a pilot framework that unifies dynamic self-debate within a single architecture. We articulate our contributions by identifying three critical challenges in adapting MoE for debate and presenting our corresponding architectural advantages:

    \noindent $\bullet$ \textbf{Dual-Routing for Dialectical Flow Control.}  Standard MoE architectures typically rely on a single router to dispatch tokens to experts. This design is insufficient for debate, which requires orchestration across different dimensions: assigning specific \emph{roles} (e.g., proposer vs. critic) and managing the \emph{process flow} (e.g., debating vs. concluding). A single router cannot simultaneously optimize for these orthogonal objectives. We propose a dual-routing mechanism that decouples role allocation from stage management. By introducing specialized gating functions, our model dynamically regulates when to debate (invoking dialectical confrontation) and when to summarize (invoking synthesis). This allows the system to switch strategies adaptively, skipping unnecessary argumentation for simple sub-problems or invoking early synthesis to resolve conflicts, thereby optimizing the reasoning trajectory.
    
    \noindent $\bullet$ \textbf{Momentum Switching for Consistency.}
Conventional MoE gating operates at the token level and is sensitive to local noise and minor representation fluctuations.
In a debate context, such jittery routing may swap experts too frequently (e.g., flipping between a ``pro'' and a ``critic'' stance within a short span), fragmenting the reasoning trace and hurting coherence.
We introduce \emph{Momentum Switching}, which smooths routing decisions via local contextual aggregation so expert usage evolves more gradually over time, improving short-range persistency while remaining responsive to genuine context shifts.
This stabilization is lightweight, autoregressive-friendly, and complements our dual-routing design by reducing unnecessary role flips during multi-turn reasoning.
    
    \noindent $\bullet$ \textbf{Unified Efficient Self-Debate.} Prior multi-agent debate frameworks rely on instantiating multiple distinct models or maintaining several heavy copies of LLMs to simulate interaction. This inter-model communication is computationally expensive, memory-intensive, and introduces significant latency due to network overhead or context switching. We propose a unified self-debate mode based on the MoE architecture. By encapsulating diverse debating personas into lightweight, specialized expert modules within a \emph{single} backbone, we eliminate the need for external multi-model orchestration. This design reduces inter-agent communication cost while retaining the behavioral diversity required for effective self-debate, allowing for efficient, high-quality reasoning within a compact computational budget.

\begin{figure*}
    \centering
    \includegraphics[width=1.0\textwidth]{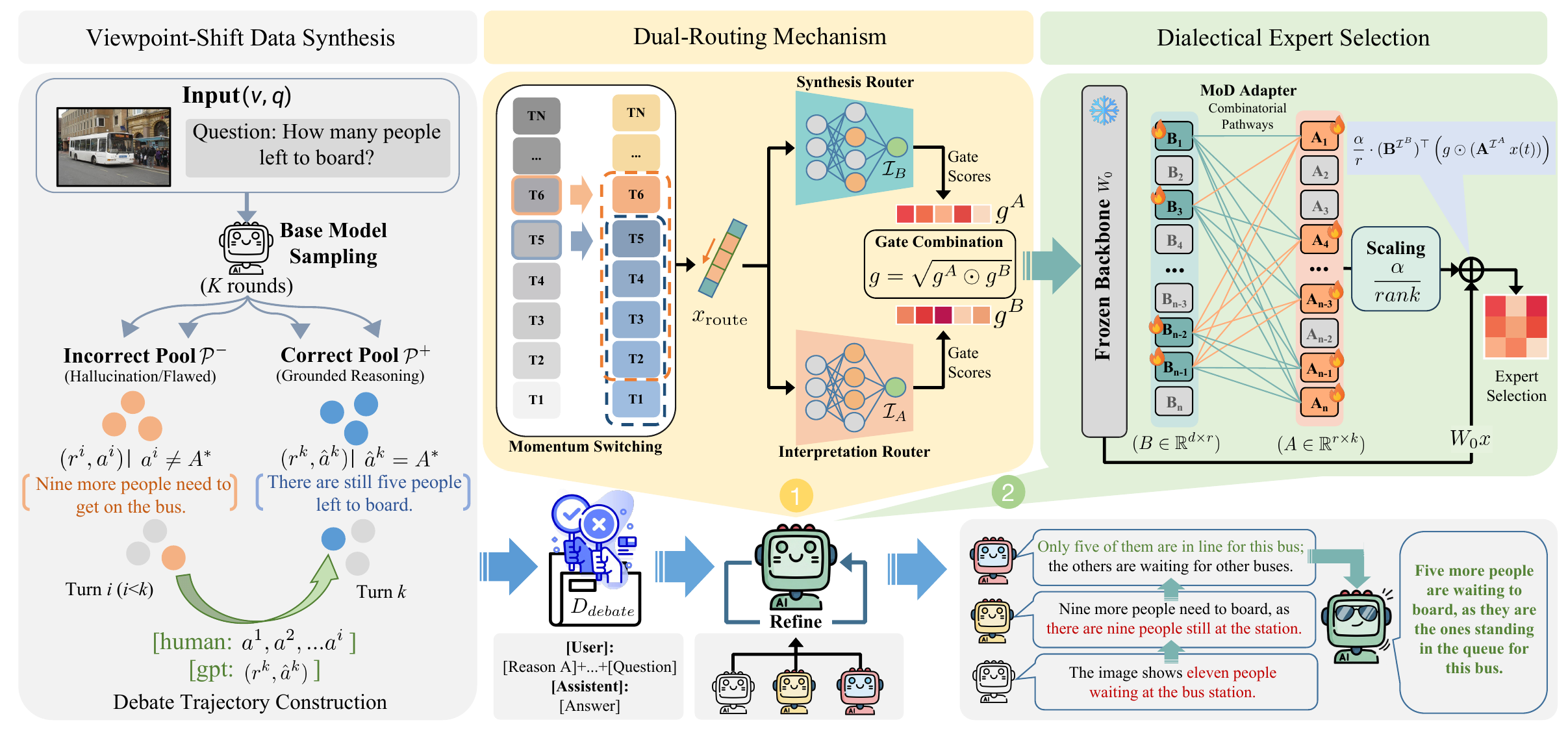}
    \Description{Overview of Mixture-of-Debaters (MoD). Left: Viewpoint-Shift Data Synthesis constructs debate trajectories from correct and incorrect samples for belief revision. Right: MoD architecture with dual-routing, momentum switching, and decoupled A-side/B-side expert pools for diverse reasoning pathways.}
    \caption{Overview of Mixture-of-Debaters (MoD). Left: Viewpoint-Shift Data Synthesis constructs debate trajectories from correct and incorrect samples for belief revision. Right: MoD architecture with dual-routing, momentum switching, and decoupled A-side/B-side expert pools for diverse reasoning pathways.}
    \label{fig:model}
\end{figure*}

\section{Related Work}
\label{sec:related}

\noindent\textbf{Multi-Agent Reasoning and Debate.}
Multi-agent debate has emerged as an effective paradigm for improving reasoning by exposing latent inconsistencies through adversarial critique and cross-examination~\cite{zhang2025madawsd,liang2024encouraging,bo2024reflective}. 
Recent studies extend this idea to multimodal settings, showing that interaction among multiple vision-language agents can improve reasoning reliability and reduce hallucination~\cite{yu2024mitigating,liang2025multi,liang2026multi, hu2025removal,zheng2025learning}. 
This line is further generalized by Mixture-of-Agents (MoA), which aggregates responses from multiple external models through layered coordination~\cite{wang2024mixture}. 
Despite their effectiveness, these approaches typically require multiple model instances or external APIs as independent agents, leading to substantial memory, communication, and latency overhead. 
Moreover, their performance often depends on the zero-shot debate capability of the underlying models, which may limit gains when base models are weak or prone to shallow contradiction and paraphrastic disagreement~\cite{sub2025multiagent,choi2025debate}.

\noindent\textbf{Internalized Deliberation and Self-Correction.}
To reduce the cost of explicit multi-agent orchestration, a parallel line of work seeks to internalize deliberative behavior within a single model. Early efforts such as Debate, Reflect, and Distill (DRD) distill multi-agent debate trajectories into a student model~\cite{zhou2025debate}, while SMoA emulates MoA-style aggregation using a single LLM that generates multiple candidate responses~\cite{li2025smoa}. More recent work has increasingly shifted toward strengthening self-critique and self-correction directly. For example, S$^2$R teaches models to self-verify and self-correct during inference through lightweight supervision and reinforcement learning~\cite{ma2025s2r}, SCRIT improves critique ability using self-generated training data~\cite{tang2025self}, and SPOC enables interleaved solution generation and verification within a single inference pass~\cite{zhao2025boosting}. In multimodal reasoning, Sherlock further extends this line by introducing trajectory-level self-correction for vision-language models~\cite{ding2025sherlock}. At the same time, recent evaluations such as RealCritic and Self-Correction Bench suggest that self-critique remains unreliable for many current models, especially in self-critique and iterative correction settings~\cite{tang2025realcritic,tsui2025self}. Overall, these methods internalize deliberation, but still rely on explicit critique loops, multi-pass generation, or specialized self-improvement procedures. This motivates architectures that support deliberation more directly and efficiently within a single forward reasoning framework.

\noindent\textbf{Parameter-Efficient Mixture-of-Experts.}
Mixture-of-Experts (MoE) provides a natural mechanism for conditional computation within a single backbone, making it a promising foundation for scalable reasoning models~\cite{GShard2021,zhou2022mixture}. 
Recent parameter-efficient variants combine MoE with low-rank adaptation by replacing dense adaptation with sparsely activated LoRA experts. 
Early approaches such as MoE-LoRA, AdaMV-MoE, MixLoRA, and Uni-MoE show that routing among lightweight adapters can improve the capacity--efficiency trade-off in both language and multimodal settings~\cite{liu2023moelora,li2024mixlora, chen2023adamv,li2025uni}. 
More recent work further advances this line through improved optimization, layer-wise expert allocation, expert specialization, and adaptive expert selection~\cite{sun2025stronger,gao2025mola,feng2025comoe,kunwar2025tt}. 
However, most existing PE-MoE methods still inherit two assumptions that are not ideal for debate-style reasoning. 
First, each expert is typically instantiated as a monolithic adapter unit, coupling the down- and up-projection under a single routing decision. This limits the asymmetric routing needed for dialectical reasoning. 
Second, prior PE-MoE methods predominantly adopt token-level routing~\cite{li2024mixlora, zhang2025more}. While effective for general sequence modeling, such fine-grained routing may cause unstable expert assignments across neighboring tokens, making it less suitable for dialectical generation requiring local coherence.

In contrast to the above literature, our Mixture-of-Debaters (MoD) treats debate as an internal routing problem within a single model rather than an interaction among external agents. 
MoD combines explicit viewpoint-shift supervision with a parameter-efficient MoE architecture, decouples interpretation and synthesis expert pools to enable asymmetric expert composition, and introduces dual routing together with Momentum Switching to stabilize local reasoning trajectories. 
This design preserves the behavioral diversity of debate while avoiding the overhead of conventional multi-agent systems.

\section{Mixture-of-Debaters}
\label{sec:method}

We present \textbf{Mixture-of-Debaters} (\textbf{MoD}), a unified framework that enables dynamic self-debate within a single model.
As illustrated in Figure~\ref{fig:model}, MoD builds upon a frozen vision-language backbone and introduces lightweight MoD adapters that replace standard LoRA modules.
We first describe the model architecture in \S\ref{sec:architecture}. We then present how we construct the MoD instruction-tuning dataset in \S\ref{sec:dataset}, which provides explicit supervision for viewpoint shifts. Finally, we detail the tuning procedure in \S\ref{sec:tuning}.

\subsection{Model Architecture}
\label{sec:architecture}

MoD supports both text-only and multimodal inputs, so we write the input as $(V, Q)$ with the text-only case as $V=\varnothing$. We focus on the multimodal setting where modality interactions make routing stability and perspective diversity especially valuable \cite{liang2025seeing}.

Given an image $V$ and question $Q$, the vision encoder extracts features, projects them into the language space, and concatenates them with text embeddings to form the input sequence, which is processed by Transformer blocks with MSA~\cite{vaswani2017attention} and FFN. We freeze the backbone projections (q/k/v/o) and inject MoD adapters in parallel for parameter-efficient adaptation. While standard LoRA uses a single low-rank bypass $h = W_0 x + \frac{\alpha}{r} BAx$~\cite{hu2022lora}, MoD replaces $(A,B)$ with decoupled expert pools and dynamic routing, enabling diverse perspectives within one forward pass. For each token $x^{(t)}$, the MoD adapter computes its output through three stages:

\subsubsection{Stage 1: Momentum Switching}

Conventional MoE gating operates at the token level, potentially selecting different experts for every generated token.
In a debate context, this high-frequency switching is detrimental: it leads to disjointed arguments where a model's stance fluctuates mid-sentence, destroying the logical consistency required for coherent argumentation.
We introduce a momentum switching strategy that smooths routing decisions and improves short-range expert persistency.

Token-level switching ($x_{\text{route}} = x$) is flexible but prone to noisy expert flips across adjacent tokens, fragmenting locally coherent reasoning.
Region-level switching partitions the sequence into fixed-size blocks, but rigid boundaries can cause abrupt transitions.

To maintain argumentative consistency while preserving adaptability, we adopt sliding window switching.
Rather than per-token switching, we compute a causal moving average over the most recent $\mathcal{W}$ tokens:
\begin{equation}
    x_{route}(t) = \frac{\sum_{k=\max(0, \, t-\mathcal{W}+1)}^{t} x(k)}{\min(t+1, \mathcal{W})},
\end{equation}

Where $t$ denotes the index of the current token within the batch. This mechanism smooths routing inputs over local context, reducing noisy expert switching and stabilizing expert usage across neighboring tokens.

To ensure that the routing behavior during inference aligns strictly with the training phase, we augment the conventional KV cache with a dedicated token cache with a capacity of $\mathcal{W}-1$ to each MoD layer, which is utilized to maintain the most recent tokens.

The sliding window routing and buffer design ensure compatibility with autoregressive generation, and we implement sliding windows via cumulative sums.
We set $\mathcal{W} = 16$ by default, providing sufficient context for stable routing while remaining responsive to topic shifts.

\subsubsection{Stage 2: Dual-Routing Mechanism}

Standard MoE architectures rely on a single router to dispatch tokens to experts~\cite{zhou2022mixture}.
However, effective debate requires orchestration across two orthogonal dimensions: assigning \emph{roles} (e.g., proposer vs.\ critic) and managing \emph{process flow} (e.g., debating vs.\ concluding).
A single router cannot simultaneously optimize for these objectives.
We introduce a dual-routing mechanism that decouples role allocation from stage management.

Specifically, we employ two separate routers, A and B, each parameterized by a learnable projection $W^A_g, W^B_g \in \mathbb{R}^{E \times K}$, where $E$ denotes the number of experts. 
For an input token representation $x_{\text{route}}$, the routing logits $\ell$ and selected expert indices $\mathcal{I}$ are computed as:
\begin{align}
    \ell^A &= W^A_g x_{\text{route}}, \quad \ell^B = W^B_g x_{\text{route}}, \\
    \mathcal{I}^A &= \text{TopK}(\ell^A, r), \quad \mathcal{I}^B = \text{TopK}(\ell^B, r).
\end{align}
During training, optional Gaussian noise can be added to the logits to encourage exploration.
The normalized gating scores $g^A \in \mathbb{R}^r$ and $g^B \in \mathbb{R}^r$ are obtained by applying softmax over $\ell^A$ and $\ell^B$ then normalize only on the top-$r$ selected experts:
\begin{equation}
    g^A = \left[ \frac{s^A_i}{\sum_{j \in \mathcal{I}^A} s^A_j} \right]_{i \in \mathcal{I}^A},
\end{equation}
where $s^A = \text{Softmax}(\ell^A)$, and analogously for $g^B$. Since the A-side and B-side experts are independently selected, we combine their gating scores to obtain the final weight $g$:
\begin{equation}
    g = \sqrt{g^A \odot g^B}.
\end{equation}
By default, we adopt a sqrt-product combination strategy, which captures the joint agreement of both routers while mitigating the score suppression that arises from direct multiplication.

Enabled by the decoupled design of the two routers, the model can dynamically decide when to debate and when to synthesize. For simple sub-problems, it can skip unnecessary argumentation; conversely, when conflicts emerge, it selectively allocates capacity to focus on correction.

\subsubsection{Stage 3: Dialectical Expert Pools}

We maintain two decoupled pools of rank-1 experts: interpretation experts $\mathcal{E}^A = \{A_1, \dots, A_E\}$ with $A_i \in \mathbb{R}^{1 \times k}$, and synthesis experts $\mathcal{E}^B = \{B_1, \dots, B_E\}$ with $B_i \in \mathbb{R}^{1 \times d}$.
Given selected indices $\mathcal{I}_A, \mathcal{I}_B$ and gating weight $g$, the adapter output is:
\begin{equation}
    h^{\text{MoD}}(t) = \frac{\alpha}{r} \cdot (\mathbf{B}^{\mathcal{I}^B})^\top \left( g \odot (\mathbf{A}^{\mathcal{I}^A} \, x(t)) \right),
    \label{eq:mod_forward}
\end{equation}
where $\mathbf{A}^{\mathcal{I}^A} \in \mathbb{R}^{r \times k}$ and $\mathbf{B}^{\mathcal{I}^B} \in \mathbb{R}^{r \times d}$ are stacked selected experts.
The final hidden state is $h{(t)} = W_0 x{(t)} + h^{\text{MoD}}{(t)}$.

Unlike MoE-LoRA which couples $(A_i, B_i)$ as atomic units~\cite{liu2023moelora}, our decoupled design enables $N \times N$ combinatorial pathways.
This allows the model to pair different interpretation and synthesis experts, effectively simulating diverse debating perspectives within a single forward pass.


\begin{table*}
\centering
\caption{Performance comparison on multimodal reasoning benchmarks. MoD-Single denotes single-round inference, while MoD-Debate employs multi-turn dialectical reasoning. Best results are in \textbf{bold}, and second-best are \underline{underlined}.}
\label{tab:main_results}
\begin{tabular}{l|c|cccc|cc}
\toprule
\textbf{Category} & \textbf{Text} & \multicolumn{6}{c}{\textbf{Multimodal}} \\
\midrule
\textbf{Method} & \textbf{MMLU} & \textbf{SQA/val} & \textbf{SQA/test} & \textbf{MMMU/val} & \textbf{MMStar} & \textbf{POPE} & \textbf{MME} \\
\midrule
InstructBLIP-7B~\cite{dai2023instructblip} & - & 54.7 & 54.1 & 30.6 & 32.7 & 86.1 & 1137/254 \\
Qwen-VL-Chat~\cite{bai2023qwen} & 50.7 & 65.5 & 68.8 & 37.0 & 37.5 & 61.8 & 1487/360 \\
ShareGPT4V-13B~\cite{chen2024sharegpt4v} & - & 70.7 & 72.6 & 36.6 & 38.3 & 87.5 & 1569/284 \\
LLaVA-Next-Mistral-7B~\cite{liu2024llavanext} & - & 69.5 & 73.0 & 37.0 & 38.4  & 87.3& 1512/308 \\
Gemma3-4B~\cite{team2025gemma} & - & 76.0 & 77.1 & 47.3 & 47.9 & 84.6 & 1353/391 \\
MoE-LLaVA-1.8B×4~\cite{lin2026moe} & - & - & 63.1 & - & -  & 87.0& 1291 \\
MoE-LLaVA-2.7Bx4~\cite{lin2026moe} & - & - & 68.5 & - & -  & 86.3& 1423 \\
\midrule
LLaVA-v1.6-13b~\cite{liu2024llavanext}  & 55.42 & 70.65 & 73.62 & 35.00 & 41.13  & 86.20& 1567/321 \\
w/Self-Correction~\cite{he2025self}  & \underline{56.06} & 70.98 & 74.16  & 35.66 & 41.76  & 86.91& 1564/326 \\
w/Multi-Agent Debate~\cite{liang2024encouraging}  & 55.97 & 71.23 & 74.31  & 36.22 & 42.78  & 87.27& 1567/334 \\
w/MoE-LoRA~\cite{li2024mixlora} & 55.62 & 71.56 & 74.51 & 37.29 & \underline{43.35} & 87.42 & 1564/332 \\
w/MoD-Single Round (ours) & 55.95 & \underline{71.67} & \underline{74.51}  & \underline{37.88} & 42.80 & \underline{87.51} & \textbf{1559/347} \\
w/MoD-Multi Round (ours) & \textbf{56.35} & \textbf{72.10} & \textbf{75.21} & \textbf{38.44} & \textbf{44.40} & \textbf{87.65}& \underline{1560/341} \\
\midrule
Qwen2.5VL-3b-Instruct~\cite{wang2024qwen2} & 62.84 & 79.30 & 81.40 & \textbf{53.10} & 56.30  & 85.90 & 1592/607  \\
w/Self-Correction~\cite{he2025self}  & \underline{63.70} & 79.73 & 81.57  & 48.46 & 56.57  & \textbf{86.91} & 1464/591 \\
w/Multi-Agent Debate~\cite{liang2024encouraging}  & 63.21 & 79.45 & \underline{81.62}  & 48.28 & 56.67  & \underline{86.14} & 1464/582 \\
w/MoE-LoRA~\cite{li2024mixlora} & 63.59 & 79.64 & 80.45 & 47.97 & 56.85 & 85.49 & 1592/608 \\
w/MoD-Single Round (ours) & \textbf{63.73} & \underline{79.87} & 80.66 & 46.44 & \textbf{57.84} & 85.88& \textbf{1598/623} \\
w/MoD-Multi Round (ours) & 63.69 & \textbf{79.92} & \textbf{81.71} & \underline{49.47} & \underline{57.47} & 84.16& \underline{1595/608} \\
\bottomrule
\end{tabular}
\end{table*}

\subsection{Viewpoint-Shift Data Synthesis}
\label{sec:dataset}

The architecture provides the capacity for self-debate, but realizing this potential requires supervision capturing stance-taking, counter-argumentation, and belief revision.
We synthesize training data targeting \textbf{viewpoint-shift episodes}.

Given a reasoning instance $(V, Q, A^*)$, we perform $K$ sampling rounds from a base model, partitioning responses into correct pool $\mathcal{P}^{+}$ and incorrect pool $\mathcal{P}^{-}$. We construct a scalable dataset of single-turn interactions (specifically, Human-GPT pairs). To ensure the accuracy of the final output, the target response $a'$ (assigned to the GPT role) is consistently sampled from $\mathcal{P}^{+}$. For the input prompt (assigned to the Human role), we aggregate the original question $Q$ and images $V$ with $R$ responses $(a^1, a^2, \dots, a^R)$ randomly sampled from $\mathcal{P}^{+} \cup \mathcal{P}^{-}$, supplemented by a specific debate prompt $\mathcal{D}$. 

This process is designed to simulate the model's ability to derive the correct conclusion after evaluating diverse viewpoints from other agents. Subsequently, we divided the constructed data into three categories: $\mathcal{T}_{pos}$ (consistent correct chains), $\mathcal{T}_{rev}$ (error identification and viewpoint shifts), and $\mathcal{T}_{rob}$ (maintaining correct stances despite misleading information). Finally, the fine-tuning dataset is constructed by combining data from the three categories according to different strategies.

For a specific fine-tuning data with $R$ intermediate responses $(a^1, a^2, \dots, a^R)$, the model receives, question $Q$, images $V$, and all intermediate responses concatenated with debate prompt $\mathcal{D}$ as input prompt, then the model is supervised to generate the target response $a'$.
This teaches the model to evaluate prior reasoning, identify potential errors, and produce a grounded answer that either confirms or revises earlier viewpoints.

\begin{figure*}
    \centering
    \includegraphics[width=\linewidth]{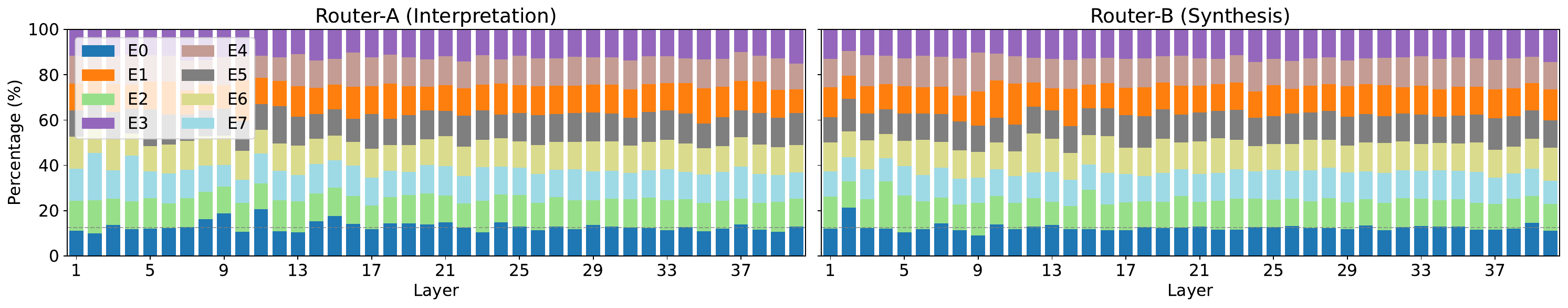}
    \Description{Comparison of layer-wise expert activation distributions between Router A and B. Each stacked bar represents the percentage of tokens routed to different experts at a given attention layer. The left panel shows Router-A (Interpretation) and the right panel shows Router-B (Synthesis), enabling a direct comparison of expert utilization patterns across layers.}
    \caption{Comparison of layer-wise expert activation distributions between Router A and B. Each stacked bar represents the percentage of tokens routed to different experts at a given attention layer. The left panel shows Router-A (Interpretation) and the right panel shows Router-B (Synthesis), enabling a direct comparison of expert utilization patterns across layers.}
    \label{fig:expert_multi}
\end{figure*}

\subsection{Training Objective}
\label{sec:tuning}

The training objective combines autoregressive language modeling loss with auxiliary load-balancing loss:
\begin{equation}
    \mathcal{L} = \mathcal{L}_{\text{reg}} + \lambda \mathcal{L}_{\text{aux}}.
\end{equation}
The autoregressive loss $\mathcal{L}_{\text{reg}}$ is computed over the target response $a'$:
\begin{equation}
    \mathcal{L}_{\text{reg}} = -\sum_{t=1}^{|a'|} \log p(a'_{t} \mid V, Q, \mathcal{D}, a^1, a^2\dots, a^R, a'_{<t}),
\end{equation}
where the model is conditioned on the visual input $V$, question $Q$, and concatenated preceding responses $(\mathcal{D}, a^1, a^2, \dots, a^R)$ as context.
The auxiliary loss $\mathcal{L}_{\text{aux}}$ encourages balanced expert utilization:
\begin{equation}
\mathcal{L}_{\text{aux}} = \frac{E}{2L} \sum_{l=1}^{L} \sum_{i=1}^{E} \left( f^A_i(l) \cdot P^A_i(l) + f^B_i(l) \cdot P^B_i(l) \right),
\end{equation}

where $E$ represents the number of experts, and $L$ is the total number of MoD layers.
The variable $f^A_i$ measures the \textit{expert usage frequency} in LoRA-A (the fraction of tokens assigning expert $i$ via TopK selection), while $P^A_i$ represents the \textit{average routing probability} for expert $i$ in LoRA-A. The definitions of $f_i^{B}$ and $P_i^{B}$ for LoRA-B follow analogously:
\begin{equation}
f^A_i = \frac{1}{T} \sum_{t=1}^{T} 1\left( i \in \mathcal{I}^A(t) \right), P^A_i = \frac{1}{T} \sum_{t=1}^{T}\ell^A_i{(t)},
\end{equation}
where $T$ denotes the total number of tokens in the batch. The total auxiliary loss $\mathcal{L}_{\text{aux}}$ is computed by averaging the load balancing objective across all layers, independently minimizing the expert collapse for both LoRA-A and LoRA-B routers. During training, the model employs self-generated content as ground truth, yielding a relatively low loss magnitude. To prevent the auxiliary loss from overshadowing the language modeling loss, we modify the conventional multi-layer auxiliary loss calculation by normalizing by the number of layers $L$, and set $\lambda = 0.01$ by default.

\section{Experiments}
\label{sec:experiments}

We conduct extensive experiments to evaluate the proposed Mixture-of-Debaters (MoD) framework. Our evaluation focuses on whether MoD improves reasoning performance over strong baselines, how each proposed component contributes to the overall gains, and whether it achieves a better accuracy-efficiency trade-off across diverse reasoning scenarios.

\subsection{Experimental Setup}
\label{sec:setup}

We build MoD upon LLaVA-v1.6-Vicuna-13B~\cite{liu2024llavanext} and Qwen 2.5VL-3B-Instruct~\cite{wang2024qwen2}, keeping base model weights frozen while training only the expert modules and routers.
We evaluate on six benchmarks: MMLU~\cite{hendryckstest2021} for multitask language understanding, ScienceQA~\cite{lu2022learn} for multimodal science reasoning, MMMU~\cite{yue2024mmmu} for college-level reasoning, MMStar~\cite{chen2024we} for vision-indispensable multimodal evaluation, POPE~\cite{lievaluating} for hallucination evaluation, and MME~\cite{fu2025mme} for perception and cognition abilities.
For MoD, we evaluate two modes: MoD-Single (single-round inference) and MoD-Multi (two-round self-debate).
Detailed implementation, configurations, and data construction details are provided in the Appendix.

\begin{figure}[t]
    \centering
    \includegraphics[width=0.9\columnwidth]{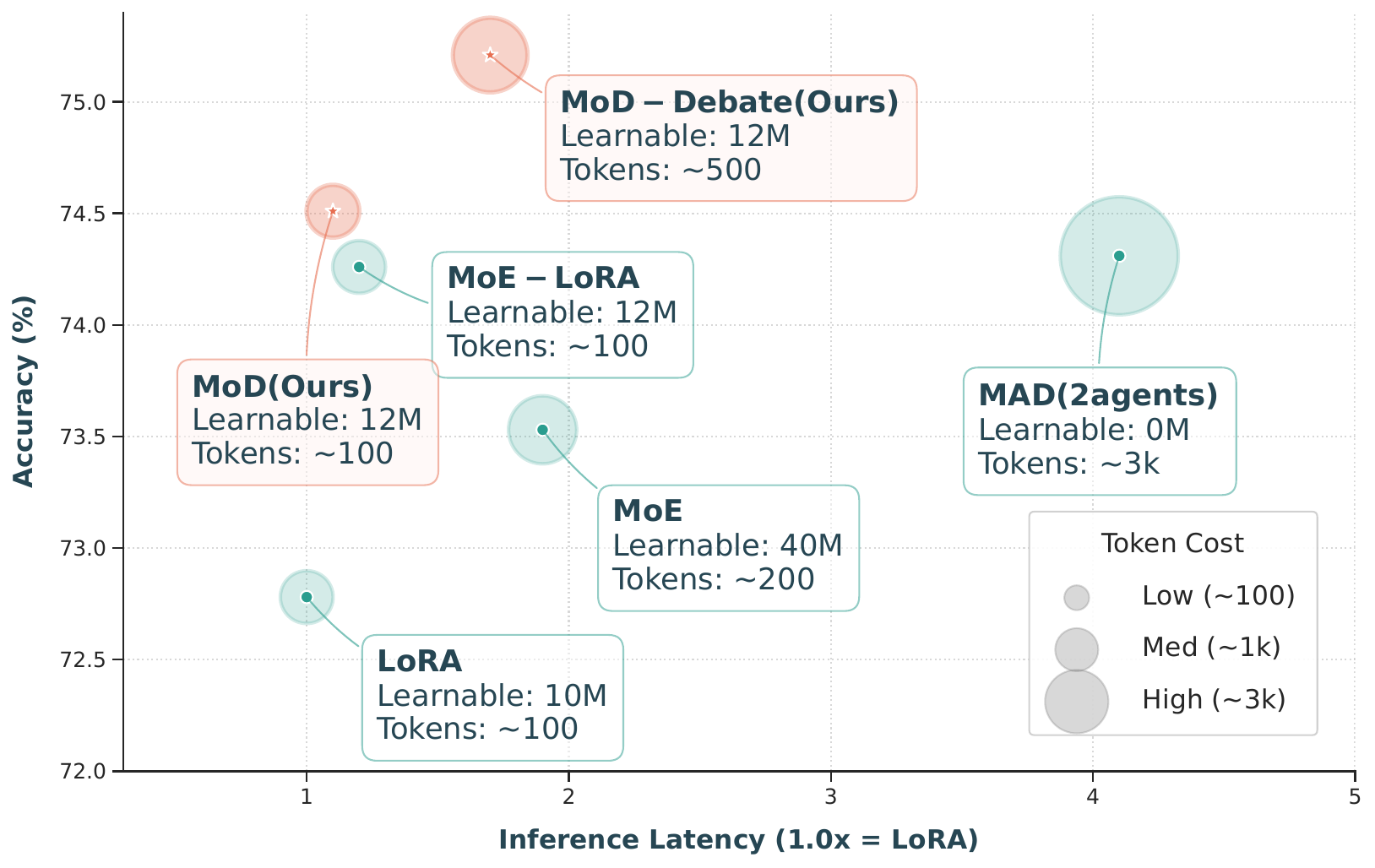}
    \Description{Efficiency comparison on ScienceQA-TEST. X-axis shows inference latency relative to LoRA (1.0$\times$). Y-axis shows accuracy. Bubble size indicates token consumption.}
    \caption{Efficiency comparison on ScienceQA-TEST. X-axis shows inference latency relative to LoRA (1.0$\times$). Y-axis shows accuracy. Bubble size indicates token consumption.}
    \label{fig:efficiency}
\end{figure}

\subsection{Main Results}
\label{sec:main_results}

As shown in Table~\ref{tab:main_results}, MoD achieves the strongest overall performance, with especially clear gains over Standard MoE-LoRA on reasoning-intensive benchmarks. On LLaVA-v1.6-13B, MoD-Multi improves ScienceQA-TEST from 74.51 to 75.21, MMMU-VAL from 37.29 to 38.44, and MMStar from 43.35 to 44.40. These results show that the proposed decoupled routing design yields consistent gains over existing routed LoRA baselines, rather than merely matching them with a different formulation. 

MoD also generalizes across backbones. On Qwen2.5VL-3B-Instruct, MoD-Multi improves ScienceQA-TEST from 80.45 to 81.71 and MMMU-VAL from 47.97 to 49.47 over Standard MoE-LoRA, while MoD-Single achieves the best MMStar score of 57.84. Overall, these results suggest that internalizing debate through dynamic expert routing provides a stronger reasoning-oriented alternative to standard MoE-LoRA-style routing.

\begin{figure}[t]
    \centering
    \includegraphics[width=1.0\columnwidth]{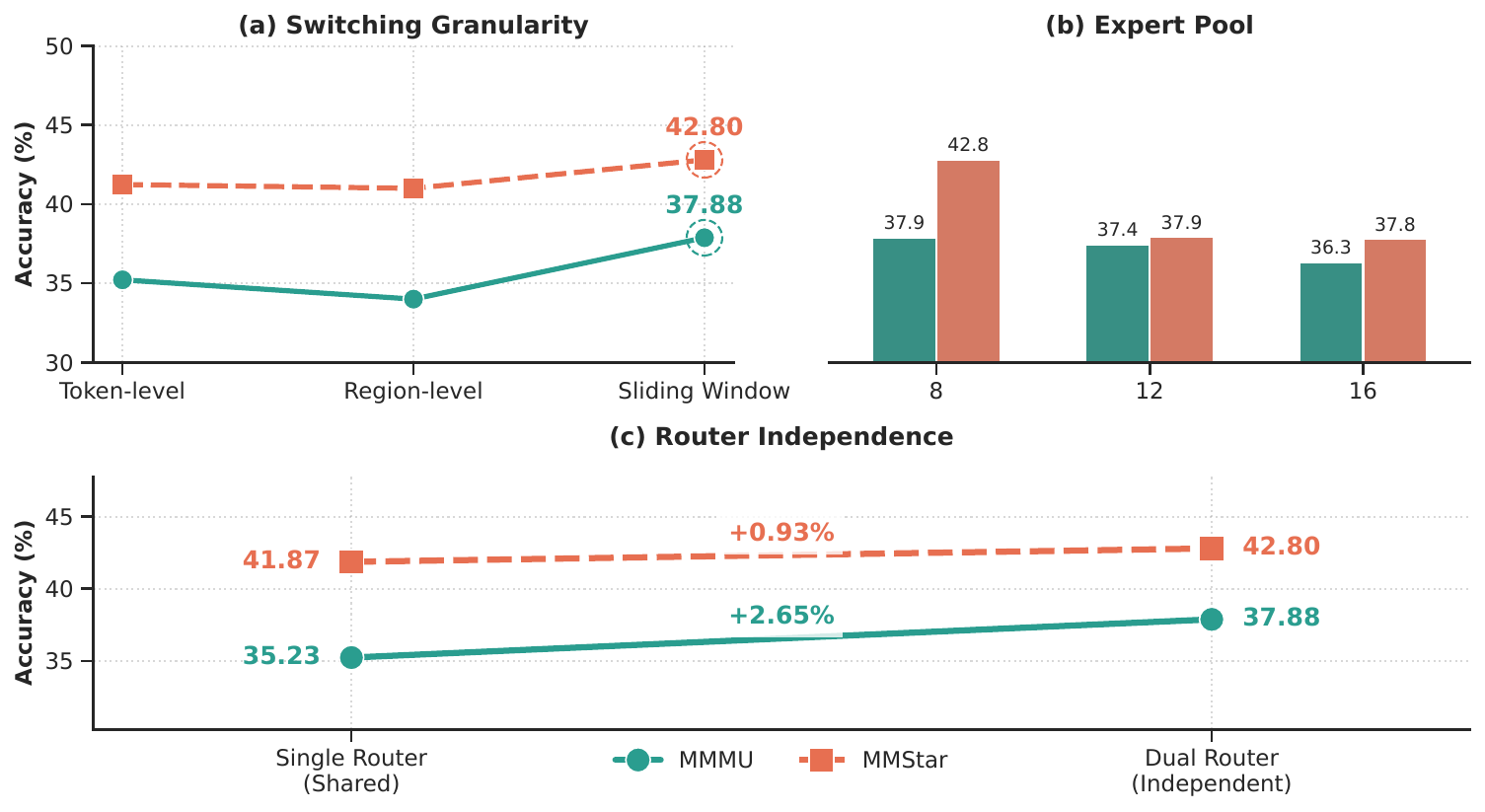}
    \Description{Ablation analysis on (a) switching granularity, (b) gating combination, and (c) router independence, validating our proposed momentum switching, sqrt-product combination, and dual-routing designs.}
    \caption{Ablation analysis on (a) switching granularity, (b) gating combination, and (c) router independence, validating our proposed momentum switching, sqrt-product combination, and dual-routing designs.}
    \label{fig:ablation_routing}
\end{figure}

\subsection{Ablation Studies}
\label{sec:ablation}

We conduct systematic ablations to isolate the contribution of each proposed component. All experiments use LLaVA-v1.6-Vicuna-13B as the base model and report accuracy on ScienceQA-TEST, POPE, MMMU-VAL and MMStar.

\subsubsection{Effect of Data Synthesis Strategy}
\label{sec:ablation_data}

Table~\ref{tab:ablation_data} examines the contribution of each trajectory topology. We compare: (1) \textbf{$\mathcal{T}_{pos}$ only}, training solely on consistent correct chains; (2) \textbf{$\mathcal{T}_{pos}$ + $\mathcal{T}_{rev}$}, adding correction trajectories; and (3) \textbf{Full} (ours), combining all three topologies including robustness trajectories $\mathcal{T}_{rob}$.

Training on $\mathcal{T}_{pos}$ alone reduces to standard supervised fine-tuning, lacking epistemic conflict to teach error recognition. Adding $\mathcal{T}_{rev}$ provides viewpoint-shift supervision, yielding substantial gains by forcing the model to identify flawed reasoning and revise its stance.The full mixture further incorporates $\mathcal{T}_{rob}$, training the model to maintain correct beliefs despite intermediate misleading information, a critical capability when debating against adversarial counterparts.

\subsubsection{Effect of Architecture Design}
\label{sec:ablation_arch}

Figure~\ref{fig:efficiency} compares four representative paradigms: (1) \textbf{Standard LoRA}, a single low-rank adapter without expert decomposition; (2) \textbf{MoE}, which replaces the feed-forward network with full-size FFN experts at substantial parameter cost; (3) \textbf{MoE-LoRA}, a parameter-efficient routed adapter baseline that couples each low-rank expert under a single routing decision; and (4) \textbf{MoD} (ours), which decouples interpretation and synthesis expert pools with independent routing. We additionally include \textbf{Multi-Agent Debate (MAD)} as an external-debate reference. This comparison addresses two questions: whether MoD improves over existing routed LoRA-style architectures, and whether internalized debate is more efficient than external multi-agent debate.

\begin{table}[t]
\centering
\small
\caption{Ablation on trajectory topology composition. $\mathcal{T}_{pos}$: correct samples only. $\mathcal{T}_{rev}$: correction trajectories. $\mathcal{T}_{rob}$: robustness trajectories with interleaved samples.}
\label{tab:ablation_data}
\begin{tabular}{lccc}
\toprule
\textbf{Trajectory Composition} & \textbf{POPE} & \textbf{SQA} & \textbf{MMMU} \\
\midrule
$\mathcal{T}_{pos}$ only & 85.15 & 73.73 & 36.56 \\
$\mathcal{T}_{pos}$ + $\mathcal{T}_{rev}$ & 86.45 & 74.22 & 37.45 \\
Full ($\mathcal{T}_{pos}$ + $\mathcal{T}_{rev}$ + $\mathcal{T}_{rob}$) & \textbf{87.51} & \textbf{74.51} & \textbf{37.88} \\
\bottomrule
\end{tabular}
\end{table}

Decoupling interpretation (A-side) and synthesis (B-side) experts enables $N \times N$ combinatorial pathways, compared with the $N$ pathways of coupled MoE-LoRA, providing richer compositional capacity at the same parameter scale.

Figure~\ref{fig:efficiency} reports the accuracy--efficiency trade-off on ScienceQA-TEST. Compared with MoE-LoRA, MoD operates at the same learnable parameter scale (12M) and similar token budget, yet achieves higher accuracy, showing that the gain comes from the proposed decoupled routing design rather than increased model size or token cost. Compared with MAD, MoD achieves better accuracy with much lower latency and token consumption by internalizing deliberation within a single backbone. Overall, these results show that MoD improves over routed PEFT baselines such as MoE-LoRA, while also offering a substantially better accuracy and efficiency trade-off than external multi-agent debate.

\begin{figure}[t]
    \centering
    \includegraphics[width=1.0\columnwidth]{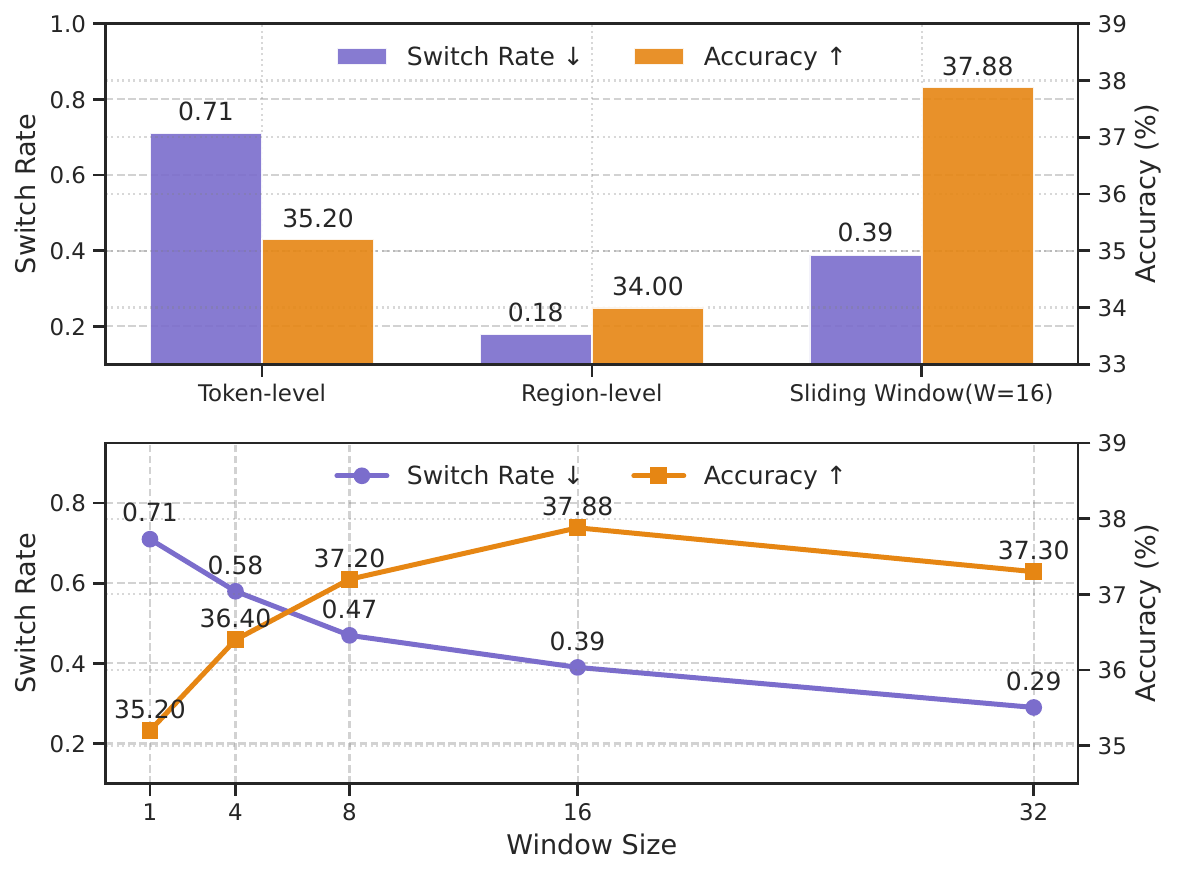}
    \Description{Routing stability analysis. \textbf{Top}: Comparison of different switching strategies in terms of switch rate and accuracy. \textbf{Bottom}: Effect of window size in Momentum Switching.}
    \caption{Routing stability analysis. \textbf{Top}: Comparison of different switching strategies in terms of switch rate and accuracy. \textbf{Bottom}: Effect of window size in Momentum Switching.}
    \label{fig:routing_stability}
\end{figure}

\subsubsection{Effect of Routing Mechanism}
\label{sec:ablation_routing}

We ablate three key design choices in our routing mechanism on MMStar and MMMU-VAL, corresponding to challenges identified in adapting MoE to dialectical reasoning.

\textbf{Dual-Routing vs. Single-Routing.}
Our first contribution argues that effective debate requires decoupling role allocation from process flow, which a single router cannot simultaneously optimize.
Figure~\ref{fig:ablation_routing}(c) validates this by comparing independent dual routers against a shared router (forcing $\mathcal{I}_A = \mathcal{I}_B$).
Dual routing achieves 42.80\% on MMStar and 37.88\% on MMMU-VAL, outperforming shared routing by +0.93\% and +2.65\% respectively.
This confirms that decoupling role allocation from process control enables asymmetric expert combinations essential for dialectical reasoning.

\textbf{Momentum Switching.}
Our second contribution addresses the instability of token-level gating, where noisy per-token decisions can trigger frequent expert switches and cause stance jitter mid-argument.
Figure~\ref{fig:ablation_routing}(a) compares three granularities: token-level achieves 35.22\% on MMMU-VAL due to erratic expert switching; region-level drops to 34.0\% as rigid boundaries disrupt autoregressive generation. Our sliding window achieves 38.44\% by stabilizing expert usage over short spans while preserving autoregressive compatibility.
On MMStar, sliding window (42.80\%) outperforms token-level (41.24\%) by 3.2\%, confirming that stable expert usage is critical for complex multimodal reasoning tasks.

\begin{figure*}
    \centering
    \includegraphics[width=0.9\linewidth]{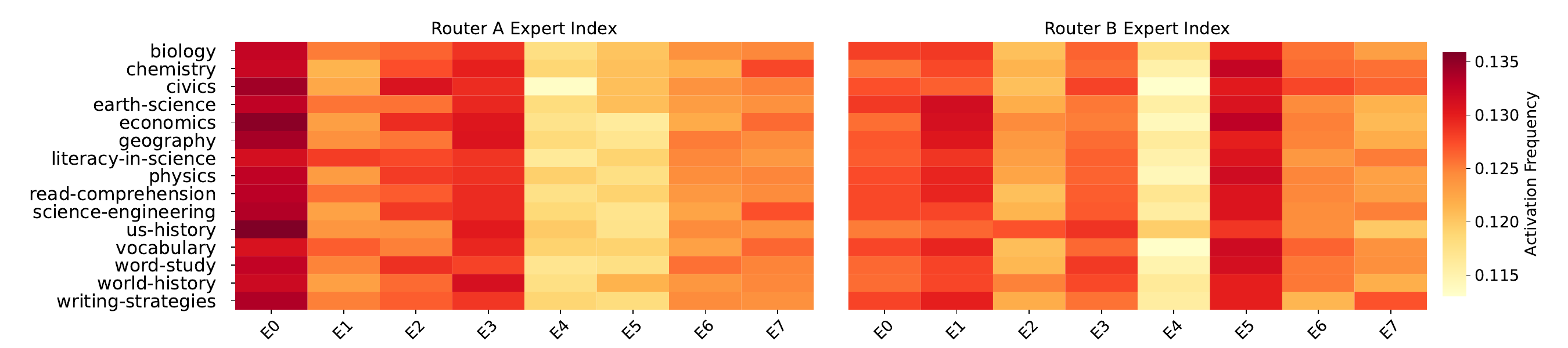}
    \Description{Category-wise expert activation heatmaps for Router A and B on the ScienceQA-TEST benchmark. Each row corresponds to a topic category and each column to an expert. Color intensity indicates the relative activation frequency of an expert within a category.}
    \caption{Category-wise expert activation heatmaps for Router A and B on the ScienceQA-TEST benchmark. Each row corresponds to a topic category and each column to an expert. Color intensity indicates the relative activation frequency of an expert within a category.}
    \label{fig:expert_heatmap}
\end{figure*}

\textbf{Expert Pool Size ($E$).}
We ablate the total number of experts in each pool (i.e., $|\mathcal{E}^A|=|\mathcal{E}^B|=E$) while fixing Top-$r=4$ for both routers.
As shown in Fig.~\ref{fig:ablation_routing}(b), $E=8$ performs best (MMMU 37.9\%, MMStar 42.8\%).
Increasing to $E=12$ reduces MMMU by 0.5 points (37.4\%) and MMStar by 4.9 points (37.9\%); $E=16$ further drops MMMU by 1.6 points (36.3\%) and MMStar by 5.0 points (37.8\%).
This suggests that enlarging the expert pool under fixed sparsity makes routing/training less effective, weakening specialization. We use $E=8$ by default.
To further verify the proposed routing design, we analyze its conflict sensitivity and routing stability.

\subsubsection{Routing Stability Analysis}
\label{sec:routing_stability}

Our second design claim is that Momentum Switching improves short-range routing consistency by reducing expert-switch jitter during autoregressive generation. We quantify this effect on the MMMU-VAL benchmark using \textbf{Switch Rate}, defined as the fraction of adjacent token pairs with different top-1 experts, where lower values indicate more stable routing. We compare three routing granularities, namely token-level routing, region-level routing, and our sliding-window Momentum Switching, and further vary the window size $\mathcal{W}$ to examine the trade-off between routing stability and task performance.

\begin{figure*}
    \centering
    \includegraphics[width=\linewidth]{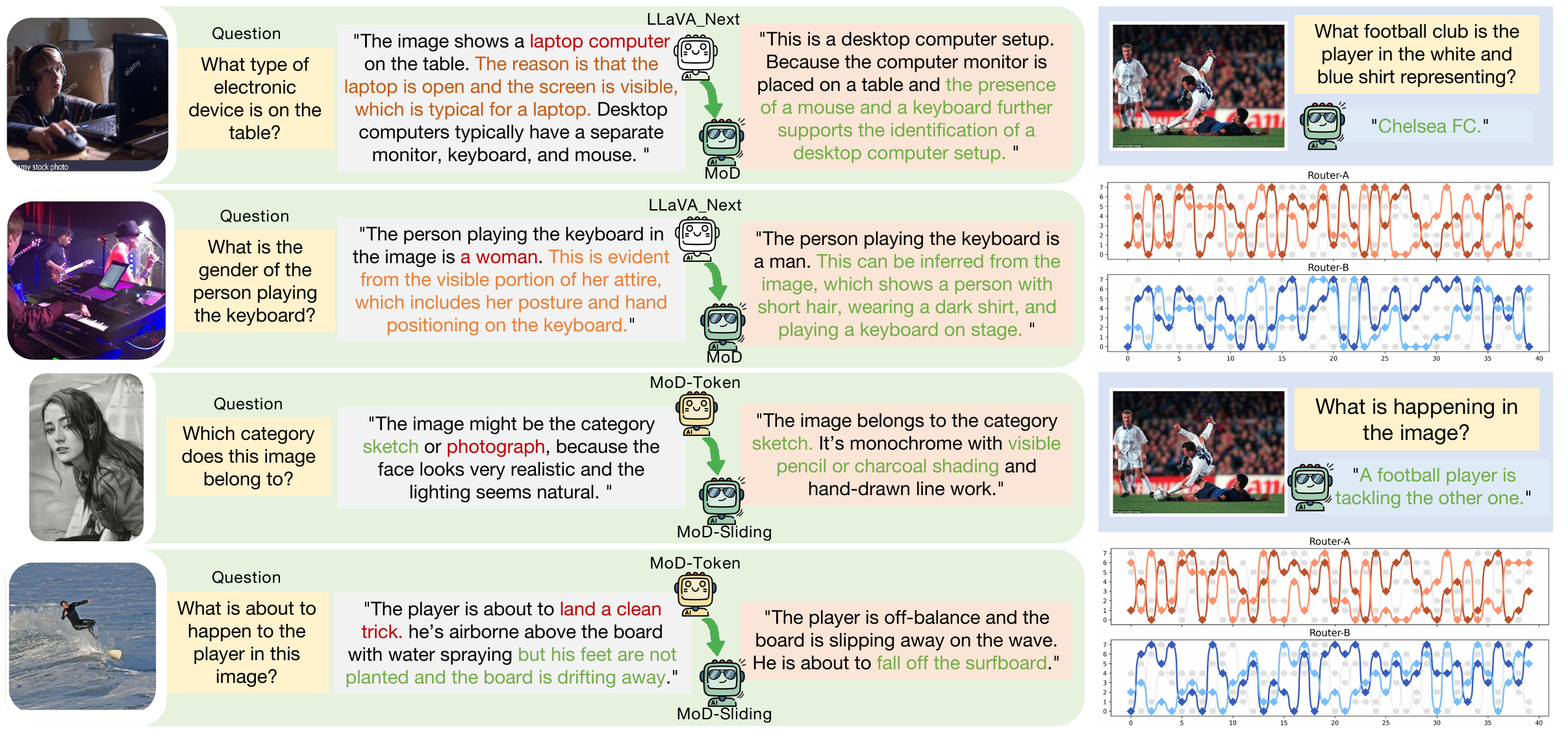}
    \Description{Case Study. \textbf{Left}: MoD produces more grounded responses than LLaVA-Next by attending to fine-grained visual details (top), and momentum switching prevents mid-response stance shifts that occur with token-level routing (bottom). \textbf{Right}: Expert activation analysis on the same image with different questions.}
    \caption{Case Study. \textbf{Left}: MoD produces more grounded responses than LLaVA-Next by attending to fine-grained visual details (top), and momentum switching prevents mid-response stance shifts that occur with token-level routing (bottom). \textbf{Right}: Expert activation analysis on the same image with different questions.}
    \label{fig:case_study}
\end{figure*}

As shown in Figure~\ref{fig:routing_stability}, the three routing strategies exhibit different stability--performance trade-offs. Token-level routing has the highest switch rate (0.71) and low accuracy (35.2\%), while region-level routing greatly reduces switching (0.18) but further lowers accuracy to 34.0\%. In contrast, sliding-window Momentum Switching achieves a lower switch rate (0.39) and the best accuracy (37.88\%), indicating improved routing stability without sacrificing reasoning flexibility.

The bottom panel shows the effect of window size $\mathcal{W}$. As $\mathcal{W}$ increases from 1 to 32, switch rate decreases from 0.71 to 0.29, confirming that larger windows smooth routing decisions. Accuracy, however, peaks at $\mathcal{W}=16$ (37.88\%) and slightly drops at $\mathcal{W}=32$ (37.3\%), indicating that overly large windows may oversmooth routing and weaken adaptability. Overall, $\mathcal{W}=16$ provides the best trade-off between routing stability and reasoning performance.

\subsubsection{Expert Utilization Analysis}
\label{sec:expert_analysis}

To validate that our architectural innovations lead to meaningful expert specialization, we analyze routing patterns across two dimensions: layer-wise activation and topic-wise distribution.

\textbf{Expert Activation Analysis.}
Figure~\ref{fig:expert_multi} compares expert activation for Router-A (role allocation) and Router-B (process control).
Both routers maintain balanced utilization across all eight experts throughout layers, confirming that our load-balancing objective prevents expert collapse.
Notably, the two routers exhibit distinct activation preferences: Router-A shows higher activation on E0, while Router-B distributes more uniformly across experts.
This divergence validates that our dual-routing mechanism learns differentiated strategies for role allocation versus process control.
Detailed activation heatmaps are provided in Appendix.

\textbf{Topic-wise Expert Specialization.}
Figure~\ref{fig:expert_heatmap} visualizes expert activation frequencies stratified by question topic for both routers.
While overall activation remains balanced, subtle topic-dependent patterns emerge.
For Router-A, experts E0 and E3 show elevated activation for science-related topics (biology, chemistry, physics), while E5 and E7 are more active for humanities topics (us-history, world-history).
Router-B exhibits complementary patterns, with E1 and E4 favored for science and E2 and E6 for humanities.
This cross-router complementarity enables the $N \times N$ combinatorial pathways central to our design: by pairing topic-specialized interpretation experts with process-appropriate synthesis experts, the model can construct diverse reasoning strategies tailored to each input.

\subsection{Case Study}
\label{sec:case_study}

Figure~\ref{fig:case_study} presents qualitative comparisons.
Firstly, compared to LLaVA-Next, MoD attends to finer-grained visual details: it correctly identifies a desktop setup by recognizing the separate monitor and keyboard, rather than misclassifying the scene as a laptop.
Secondly, the comparison between token-level routing and Momentum Switching reveals not only improved consistency, but also evidence of internal debate. Token-level routing exposes competing hypotheses, such as ``sketch or photograph'' or ``land a clean trick,'' whereas MoD-Sliding resolves them into a more coherent final judgment. This suggests that MoD internally evaluates alternative hypotheses and stabilizes them through smoother routing.
Finally, we analyze expert activations on the same image under different questions. Both routers adapt their selections according to the query, while Router-B exhibits more pronounced variation across questions, indicating stronger question-dependent synthesis behavior. In particular, the right panel shows that even for the same visual input, the routing trajectories differ substantially when the model answers ``What football club is the player representing?'' versus ``What is happening in the image?'' This supports our claim that MoD internalizes debate as dynamic coordination between interpretation and synthesis pathways, rather than a static protocol.

\section{Conclusion}
\label{sec:conclusion}

We present Mixture-of-Debaters (MoD), a framework that enables dynamic self-debate within a single model.
MoD addresses three key challenges: dual-routing decouples role allocation from process control, momentum switching maintains argumentative consistency, and unified self-debate eliminates multi-agent overhead.
Experiments show MoD outperforms both single-model baselines and multi-agent debate systems, achieving 3.7$\times$ lower latency and 87\% token reduction with only 12M additional parameters.
We hope MoD provides a practical and efficient alternative for enhancing reasoning capabilities in large language models.



\bibliographystyle{ACM-Reference-Format}
\bibliography{sample-base}

@inproceedings{he2025self,
  title={Self-correction is more than refinement: A learning framework for visual and language reasoning tasks},
  author={He, Jiayi and Lin, Hehai and Wang, Qingyun and Fung, Yi R and Ji, Heng},
  booktitle={Findings of the Association for Computational Linguistics: ACL 2025},
  pages={6405--6421},
  year={2025}
}

@inproceedings{zheng2024picture,
  title={A picture is worth a graph: A blueprint debate paradigm for multimodal reasoning},
  author={Zheng, Changmeng and Liang, Dayong and Zhang, Wengyu and Wei, Xiao-Yong and Chua, Tat-Seng and Li, Qing},
  booktitle={Proceedings of the 32nd ACM International Conference on Multimedia},
  pages={419--428},
  year={2024}
}

@book{zheng2026multimodal,
  title={Multimodal Knowledge Systems: Construction and Reasoning},
  author={Zheng, Changmeng and Li, Qing},
  year={2026},
  publisher={Springer Nature}
}

@article{liang2025seeing,
  title={Seeing Beyond the Scene: Enhancing Vision-Language Models with Interactional Reasoning},
  author={Liang, Dayong and Zheng, Changmeng and Wen, Zhiyuan and Cai, Yi and Wei, Xiao-Yong and Li, Qing},
  journal={arXiv preprint arXiv:2505.09118},
  year={2025}
}

@article{peng2025aligning,
  title={Aligning clinical needs and AI capabilities: a survey on LLMs for medical reasoning},
  author={Peng, Qi and Li, Jiatong and Huang, Sirui and Jiang, Yiyang and Gong, Kaisong and Ding, Ronger and Ye, Shijie and Wei, Xiao-Yong and Zheng, Changmeng and Li, Qing},
  year={2025},
  publisher={TechRxiv}
}

@phdthesis{zheng2025learning,
  title  = {Learning Versatile Multimodal Representation for 
            Knowledge Extraction and Reasoning},
  author = {Zheng, Changmeng},
  school = {The Hong Kong Polytechnic University},
  year   = {2025}
}

@inproceedings{liang2024encouraging,
  title={Encouraging divergent thinking in large language models through multi-agent debate},
  author={Liang, Tian and He, Zhiwei and Jiao, Wenxiang and Wang, Xing and Wang, Yan and Wang, Rui and Yang, Yujiu and Shi, Shuming and Tu, Zhaopeng},
  booktitle={Proceedings of the 2024 conference on empirical methods in natural language processing},
  pages={17889--17904},
  year={2024}
}

@inproceedings{hu2025removal,
  title={Removal of hallucination on hallucination: Debate-augmented RAG},
  author={Hu, Wentao and Zhang, Wengyu and Jiang, Yiyang and Zhang, Chen Jason and Wei, Xiaoyong and Qing, Li},
  booktitle={Proceedings of the 63rd Annual Meeting of the Association for Computational Linguistics (Volume 1: Long Papers)},
  pages={15839--15853},
  year={2025}
}

@article{wang2024mixture,
  title={Mixture-of-agents enhances large language model capabilities},
  author={Wang, Junlin and Wang, Jue and Athiwaratkun, Ben and Zhang, Ce and Zou, James},
  journal={arXiv preprint arXiv:2406.04692},
  year={2024}
}

@inproceedings{liang2026multi,
  title={Multi-Agent Undercover Gaming: Hallucination Removal Through Counterfactual Test for Multimodal Reasoning},
  author={Liang, Dayong and Wei, Xiao-Yong and Zheng, Changmeng},
  booktitle={Proceedings of the AAAI Conference on Artificial Intelligence},
  volume={40},
  number={8},
  pages={6807--6815},
  year={2026}
}

@article{tang2025self,
  title={Self-evolving critique abilities in large language models},
  author={Tang, Zhengyang and Li, Ziniu and Xiao, Zhenyang and Ding, Tian and Sun, Ruoyu and Wang, Benyou and Liu, Dayiheng and Huang, Fei and Liu, Tianyu and Yu, Bowen and others},
  journal={arXiv preprint arXiv:2501.05727},
  year={2025}
}

@article{zhao2025boosting,
  title={Boosting llm reasoning via spontaneous self-correction},
  author={Zhao, Xutong and Xu, Tengyu and Wang, Xuewei and Chen, Zhengxing and Jin, Di and Tan, Liang and Yu, Zishun and Zhao, Zhuokai and He, Yun and Wang, Sinong and others},
  journal={arXiv preprint arXiv:2506.06923},
  year={2025}
}

@article{tsui2025self,
  title={Self-correction bench: Uncovering and addressing the self-correction blind spot in large language models},
  author={Tsui, Ken},
  journal={arXiv preprint arXiv:2507.02778},
  year={2025}
}

@article{tang2025realcritic,
  title={Realcritic: Towards effectiveness-driven evaluation of language model critiques},
  author={Tang, Zhengyang and Li, Ziniu and Xiao, Zhenyang and Ding, Tian and Sun, Ruoyu and Wang, Benyou and Liu, Dayiheng and Huang, Fei and Liu, Tianyu and Yu, Bowen and others},
  journal={arXiv preprint arXiv:2501.14492},
  year={2025}
}

@inproceedings{li2025smoa,
  title={SMoA: Improving Multi-agent Large Language Models with S parse M ixture-o f-A gents},
  author={Li, Dawei and Tan, Zhen and Qian, Peijia and Li, Yifan and Chaudhary, Kumar and Hu, Lijie and Shen, Jiayi},
  booktitle={Pacific-Asia Conference on Knowledge Discovery and Data Mining},
  pages={54--65},
  year={2025},
}

@article{ding2025sherlock,
  title={Sherlock: Self-correcting reasoning in vision-language models},
  author={Ding, Yi and Zhang, Ruqi},
  journal={arXiv preprint arXiv:2505.22651},
  year={2025}
}

@inproceedings{ma2025s2r,
  title={S2r: Teaching llms to self-verify and self-correct via reinforcement learning},
  author={Ma, Ruotian and Wang, Peisong and Liu, Cheng and Liu, Xingyan and Chen, Jiaqi and Zhang, Bang and Zhou, Xin and Du, Nan and Li, Jia},
  booktitle={Proceedings of the 63rd Annual Meeting of the Association for Computational Linguistics (Volume 1: Long Papers)},
  pages={22632--22654},
  year={2025}
}

@inproceedings{zhou2025debate,
  title={Debate, reflect, and distill: Multi-agent feedback with tree-structured preference optimization for efficient language model enhancement},
  author={Zhou, Xiaofeng and Huang, He-Yan and Liao, Lizi},
  booktitle={Findings of the Association for Computational Linguistics: ACL 2025},
  pages={9122--9137},
  year={2025}
}

@article{hendryckstest2021,
  title={Measuring Massive Multitask Language Understanding},
  author={Dan Hendrycks and Collin Burns and Steven Basart and Andy Zou and Mantas Mazeika and Dawn Song and Jacob Steinhardt},
  journal={Proceedings of the International Conference on Learning Representations (ICLR)},
  year={2021}
}

@article{lin2026moe,
  title={Moe-llava: Mixture of experts for large vision-language models},
  author={Lin, Bin and Tang, Zhenyu and Ye, Yang and Huang, Jinfa and Zhang, Junwu and Pang, Yatian and Jin, Peng and Ning, Munan and Luo, Jiebo and Yuan, Li},
  journal={IEEE Transactions on Multimedia},
  year={2026},
  publisher={IEEE}
}

@article{hu2022lora,
  title={Lora: Low-rank adaptation of large language models.},
  author={Hu, Edward J and Shen, Yelong and Wallis, Phillip and Allen-Zhu, Zeyuan and Li, Yuanzhi and Wang, Shean and Wang, Lu and Chen, Weizhu and others},
  journal={ICLR},
  volume={1},
  number={2},
  pages={3},
  year={2022}
}

@article{vaswani2017attention,
  title={Attention is all you need},
  author={Vaswani, Ashish and Shazeer, Noam and Parmar, Niki and Uszkoreit, Jakob and Jones, Llion and Gomez, Aidan N and Kaiser, {\L}ukasz and Polosukhin, Illia},
  journal={Advances in neural information processing systems},
  volume={30},
  year={2017}
}

@article{bai2023qwen,
  title={Qwen-vl: A frontier large vision-language model with versatile abilities},
  author={Bai, Jinze and Bai, Shuai and Yang, Shusheng and Wang, Shijie and Tan, Sinan and Wang, Peng and Lin, Junyang and Zhou, Chang and Zhou, Jingren},
  journal={arXiv preprint arXiv:2308.12966},
  volume={1},
  number={2},
  pages={3},
  year={2023}
}

@article{dai2023instructblip,
  title={Instructblip: Towards general-purpose vision-language models with instruction tuning},
  author={Dai, Wenliang and Li, Junnan and Li, Dongxu and Tiong, Anthony and Zhao, Junqi and Wang, Weisheng and Li, Boyang and Fung, Pascale N and Hoi, Steven},
  journal={Advances in neural information processing systems},
  volume={36},
  pages={49250--49267},
  year={2023}
}

@inproceedings{zhang2025madawsd,
  title={MADAWSD: Multi-Agent Debate Framework for Adversarial Word Sense Disambiguation},
  author={Zhang, Kaiyuan and Liu, Qian and Zhang, Luyang and Zheng, Chaoqun and Li, Shuaimin and Xu, Bing and Yang, Muyun and Qiao, Xinxiao and Lu, Wenpeng},
  booktitle={Proceedings of the 2025 Conference on Empirical Methods in Natural Language Processing},
  pages={22294--22313},
  year={2025}
}

@article{bo2024reflective,
  title={Reflective multi-agent collaboration based on large language models},
  author={Bo, Xiaohe and Zhang, Zeyu and Dai, Quanyu and Feng, Xueyang and Wang, Lei and Li, Rui and Chen, Xu and Wen, Ji-Rong},
  journal={Advances in Neural Information Processing Systems},
  volume={37},
  pages={138595--138631},
  year={2024}
}

@inproceedings{yu2024mitigating,
  title={Mitigating Large Vision-Language Model Hallucination at Post-hoc via Multi-agent System},
  author={Yu, Chung-En Johnny and Jalaian, Brian and Bastian, Nathaniel D},
  booktitle={Proceedings of the AAAI Symposium Series},
  volume={4},
  number={1},
  pages={110--113},
  year={2024}
}

@article{liang2025multi,
  title={Multi-agent Undercover Gaming: Hallucination Removal via Counterfactual Test for Multimodal Reasoning},
  author={Liang, Dayong and Wei, Xiao-Yong and Zheng, Changmeng},
  journal={arXiv preprint arXiv:2511.11182},
  year={2025}
}

@article{didolkar2024metacognitive,
  title={Metacognitive capabilities of llms: An exploration in mathematical problem solving},
  author={Didolkar, Aniket and Goyal, Anirudh and Ke, Nan Rosemary and Guo, Siyuan and Valko, Michal and Lillicrap, Timothy and Jimenez Rezende, Danilo and Bengio, Yoshua and Mozer, Michael C and Arora, Sanjeev},
  journal={Advances in Neural Information Processing Systems},
  volume={37},
  pages={19783--19812},
  year={2024}
}

@inproceedings{seals2024evaluating,
  title={Evaluating the deductive competence of large language models},
  author={Seals, S and Shalin, Valerie},
  booktitle={Proceedings of the 2024 Conference of the North American Chapter of the Association for Computational Linguistics: Human Language Technologies (Volume 1: Long Papers)},
  pages={8614--8630},
  year={2024}
}

@article{wei2022chain,
  title={Chain-of-thought prompting elicits reasoning in large language models},
  author={Wei, Jason and Wang, Xuezhi and Schuurmans, Dale and Bosma, Maarten and Xia, Fei and Chi, Ed and Le, Quoc V and Zhou, Denny and others},
  journal={Advances in neural information processing systems},
  volume={35},
  pages={24824--24837},
  year={2022}
}

@inproceedings{lievaluating,
    title = "Evaluating Object Hallucination in Large Vision-Language Models",
    author = "Li, Yifan  and
      Du, Yifan  and
      Zhou, Kun  and
      Wang, Jinpeng  and
      Zhao, Xin  and
      Wen, Ji-Rong",
    editor = "Bouamor, Houda  and
      Pino, Juan  and
      Bali, Kalika",
    booktitle = "Proceedings of the 2023 Conference on Empirical Methods in Natural Language Processing",
    month = dec,
    year = "2023",
    address = "Singapore",
    pages = "292--305",
}

@inproceedings{fu2025mme,
  title={Mme: A comprehensive evaluation benchmark for multimodal large language models},
  author={Fu, Chaoyou and Chen, Peixian and Shen, Yunhang and Qin, Yulei and Zhang, Mengdan and Lin, Xu and Yang, Jinrui and Zheng, Xiawu and Li, Ke and Sun, Xing and others},
  booktitle={The Thirty-ninth Annual Conference on Neural Information Processing Systems Datasets and Benchmarks Track},
  year={2025}
}

@inproceedings{li2025advancing,
  title={Advancing collaborative debates with role differentiation through multi-agent reinforcement learning},
  author={Li, Haoran and Su, Ziyi and Xue, Yun and Tian, Zhiliang and Song, Yiping and Huang, Minlie},
  booktitle={Proceedings of the 63rd Annual Meeting of the Association for Computational Linguistics (Volume 1: Long Papers)},
  pages={22655--22666},
  year={2025}
}

@inproceedings{chen2023adamv,
  title={Adamv-moe: Adaptive multi-task vision mixture-of-experts},
  author={Chen, Tianlong and Chen, Xuxi and Du, Xianzhi and Rashwan, Abdullah and Yang, Fan and Chen, Huizhong and Wang, Zhangyang and Li, Yeqing},
  booktitle={Proceedings of the IEEE/CVF International Conference on Computer Vision},
  pages={17346--17357},
  year={2023}
}

@article{sub2025multiagent,
  title={Multiagent finetuning: Self improvement with diverse reasoning chains},
  author={Subramaniam, Vighnesh and Du, Yilun and Tenenbaum, Joshua B and Torralba, Antonio and Li, Shuang and Mordatch, Igor},
  journal={arXiv preprint arXiv:2501.05707},
  year={2025}
}

@inproceedings{choi2025debate,
  title={Debate or Vote: Which Yields Better Decisions in Multi-Agent Large Language Models?},
  author={Choi, Hyeong Kyu and Zhu, Xiaojin and Li, Sharon},
  booktitle={Advances in Neural Information Processing Systems},
  year={2025}
}

@inproceedings{GShard2021,
  author={Dmitry Lepikhin and HyoukJoong Lee and Yuanzhong Xu and Dehao Chen and Orhan Firat and Yanping Huang and Maxim Krikun and Noam Shazeer and Zhifeng Chen},
  title={GShard: Scaling Giant Models with Conditional Computation and Automatic Sharding},
  booktitle={9th International Conference on Learning Representations, {ICLR} 2021,Virtual Event, Austria, May 3-7, 2021},
  year={2021},
}

@article{zhou2022mixture,
  title={Mixture-of-experts with expert choice routing},
  author={Zhou, Yanqi and Lei, Tao and Liu, Hanxiao and Du, Nan and Huang, Yanping and Zhao, Vincent and Dai, Andrew M and Le, Quoc V and Laudon, James and others},
  journal={Advances in Neural Information Processing Systems},
  volume={35},
  pages={7103--7114},
  year={2022}
}

@inproceedings{liu2023llava,
author      = {Liu, Haotian and Li, Chunyuan and Wu, Qingyang and Lee, Yong Jae},
title       = {Visual Instruction Tuning},
booktitle   = {NeurIPS},
year        = {2023}
}

@article{liu2023moelora,
  title={Moelora: An moe-based parameter efficient fine-tuning method for multi-task medical applications},
  author={Liu, Qidong and Wu, Xian and Zhao, Xiangyu and Zhu, Yuanshao and Xu, Derong and Tian, Feng and Zheng, Yefeng},
  journal={CoRR},
  year={2023}
}

@misc{li2024mixlora,
      title={MixLoRA: Enhancing Large Language Models Fine-Tuning with LoRA-based Mixture of Experts}, 
      author={Dengchun Li and Yingzi Ma and Naizheng Wang and Zhengmao Ye and Zhiyuan Cheng and Yinghao Tang and Yan Zhang and Lei Duan and Jie Zuo and Cal Yang and Mingjie Tang},
      year={2024},
      eprint={2404.15159},
      archivePrefix={arXiv},
      primaryClass={cs.CL}
}

@article{li2025uni,
  title={Uni-moe: Scaling unified multimodal llms with mixture of experts},
  author={Li, Yunxin and Jiang, Shenyuan and Hu, Baotian and Wang, Longyue and Zhong, Wanqi and Luo, Wenhan and Ma, Lin and Zhang, Min},
  journal={IEEE Transactions on Pattern Analysis and Machine Intelligence},
  year={2025},
  publisher={IEEE}
}

@misc{liu2024llavanext,
    title={LLaVA-NeXT: Improved reasoning, OCR, and world knowledge},
    url={https://llava-vl.github.io/blog/2024-01-30-llava-next/},
    author={Liu, Haotian and Li, Chunyuan and Li, Yuheng and Li, Bo and Zhang, Yuanhan and Shen, Sheng and Lee, Yong Jae},
    month={January},
    year={2024}
}

@inproceedings{lu2022learn,
    title={Learn to Explain: Multimodal Reasoning via Thought Chains for Science Question Answering},
    author={Lu, Pan and Mishra, Swaroop and Xia, Tony and Qiu, Liang and Chang, Kai-Wei and Zhu, Song-Chun and Tafjord, Oyvind and Clark, Peter and Ashwin Kalyan},
    booktitle={The 36th Conference on Neural Information Processing Systems (NeurIPS)},
    year={2022}
}

@inproceedings{yue2024mmmu,
  title={Mmmu: A massive multi-discipline multimodal understanding and reasoning benchmark for expert agi},
  author={Yue, Xiang and Ni, Yuansheng and Zhang, Kai and Zheng, Tianyu and Liu, Ruoqi and Zhang, Ge and Stevens, Samuel and Jiang, Dongfu and Ren, Weiming and Sun, Yuxuan and others},
  booktitle={Proceedings of the IEEE/CVF Conference on Computer Vision and Pattern Recognition},
  pages={9556--9567},
  year={2024}
}

@article{chen2024we,
  title={Are we on the right way for evaluating large vision-language models?},
  author={Chen, Lin and Li, Jinsong and Dong, Xiaoyi and Zhang, Pan and Zang, Yuhang and Chen, Zehui and Duan, Haodong and Wang, Jiaqi and Qiao, Yu and Lin, Dahua and others},
  journal={Advances in Neural Information Processing Systems},
  volume={37},
  pages={27056--27087},
  year={2024}
}

@article{wang2024qwen2,
  title={Qwen2-vl: Enhancing vision-language model's perception of the world at any resolution},
  author={Wang, Peng and Bai, Shuai and Tan, Sinan and Wang, Shijie and Fan, Zhihao and Bai, Jinze and Chen, Keqin and Liu, Xuejing and Wang, Jialin and Ge, Wenbin and others},
  journal={arXiv preprint arXiv:2409.12191},
  year={2024}
}

@inproceedings{chen2024sharegpt4v,
  title={Sharegpt4v: Improving large multi-modal models with better captions},
  author={Chen, Lin and Li, Jinsong and Dong, Xiaoyi and Zhang, Pan and He, Conghui and Wang, Jiaqi and Zhao, Feng and Lin, Dahua},
  booktitle={European Conference on Computer Vision},
  pages={370--387},
  year={2024}
}

@article{team2025gemma,
  title={Gemma 3 technical report},
  author={Team, Gemma and Kamath, Aishwarya and Ferret, Johan and Pathak, Shreya and Vieillard, Nino and Merhej, Ramona and Perrin, Sarah and Matejovicova, Tatiana and Ram{\'e}, Alexandre and Rivi{\`e}re, Morgane and others},
  journal={arXiv preprint arXiv:2503.19786},
  year={2025}
}

@article{sun2025stronger,
  title={A stronger mixture of low-rank experts for fine-tuning foundation models},
  author={Sun, Mengyang and Wang, Yihao and Feng, Tao and Zhang, Dan and Zhu, Yifan and Tang, Jie},
  journal={arXiv preprint arXiv:2502.15828},
  year={2025}
}

@article{feng2025comoe,
  title={CoMoE: Contrastive Representation for Mixture-of-Experts in Parameter-Efficient Fine-tuning},
  author={Feng, Jinyuan and Wei, Chaopeng and Qiu, Tenghai and Hu, Tianyi and Pu, Zhiqiang},
  journal={arXiv preprint arXiv:2505.17553},
  year={2025}
}

@inproceedings{zhang2025more,
  title={More: A mixture of low-rank experts for adaptive multi-task learning},
  author={Zhang, Dacao and Zhang, Kun and Chu, Shimao and Wu, Le and Li, Xin and Wei, Si},
  booktitle={Findings of the Association for Computational Linguistics: ACL 2025},
  pages={1311--1324},
  year={2025}
}

@inproceedings{kunwar2025tt,
  title={TT-LoRA MoE: Using Parameter-Efficient Fine-Tuning and Sparse Mixture-Of-Experts},
  author={Kunwar, Pradip and Vu, Minh N and Gupta, Maanak and Abdelsalam, Mahmoud and Bhattarai, Manish},
  booktitle={Proceedings of the International Conference for High Performance Computing, Networking, Storage and Analysis},
  pages={1332--1350},
  year={2025}
}

@inproceedings{gao2025mola,
  title={MoLA: MoE LoRA with layer-wise expert allocation},
  author={Gao, Chongyang and Chen, Kezhen and Rao, Jinmeng and Liu, Ruibo and Sun, Baochen and Zhang, Yawen and Peng, Daiyi and Guo, Xiaoyuan and Subrahmanian, VS},
  booktitle={Findings of the Association for Computational Linguistics: NAACL 2025},
  pages={5097--5112},
  year={2025}
}

\appendix

\section{Algorithm}
\label{app:algorithm}

Algorithm~\ref{alg:mod} presents the complete forward pass of a specific MoD layer, including dual-routing, Momentum Switching via sliding window, and gating score combination.

\begin{algorithm}
\caption{Mixture-of-Debaters Layer Forward Pass}
\label{alg:mod}
\begin{algorithmic}[1]
\REQUIRE Input batch sequence $X = \{x(1), x(2), \dots, x(T)\}$, window size $\mathcal{W}$, top-k $r$
\REQUIRE Expert pools $\mathcal{E}^A = \{A_1, \dots, A_E\}$, $\mathcal{E}^B = \{B_1, \dots, B_E\}$
\REQUIRE Router projections $W^A_g$, $W^B_g$
\ENSURE Output sequence $H = \{h(1), h(2), \dots, h(T)\}$, $\mathcal{L}_{\text{aux\_layer}}$

\FOR{$t = 1$ to $T$}
    \STATE \textbf{// Momentum Switching via Sliding Window}
    \STATE  $x_{route}(t) = \frac{\sum_{k=\max(0, \, t-\mathcal{W}+1)}^{t} x(k)}{\min(t+1, \mathcal{W})}$
    \STATE \textbf{// Dual-Routing: Independent Expert Selection}
    \STATE $\ell^A \gets W^A_g \cdot x_{route}(t)$
    \STATE $\ell^B \gets W^B_g \cdot x_{route}(t)$
    \STATE $\mathcal{I}^A \gets \text{TopK}(\ell^A, r)$
    \STATE $\mathcal{I}^B \gets \text{TopK}(\ell^B, r)$
    \STATE $s^A \gets \text{Softmax}(\ell^A)$
    \STATE $s^B \gets \text{Softmax}(\ell^B)$
    \STATE $g^A \gets \text{Normalize}(s^A[\mathcal{I}^A])$
    \STATE $g^B \gets \text{Normalize}(s^B[\mathcal{I}^B])$
    \STATE \textbf{// Gating Score Combination}
    \STATE $g \gets \sqrt{g^A \odot g^B}$
    \STATE \textbf{// MoD Adapter Forward}
    \STATE $\mathbf{A}^{\mathcal{I}^A} \gets \text{Stack}(\{A_i\}_{i \in \mathcal{I}^A})$ \COMMENT{$\mathbb{R}^{r \times k}$}
    \STATE $\mathbf{B}^{\mathcal{I}^B} \gets \text{Stack}(\{B_i\}_{i \in \mathcal{I}^B})$ \COMMENT{$\mathbb{R}^{r \times d}$}
    \STATE $h^{\text{MoD}}(t) \gets \frac{\alpha}{r} \cdot {\mathbf{B}^{{\mathcal{I}_B}}}^\top (g \odot (\mathbf{A}^{\mathcal{I}_A} \cdot x(t)))$
    \STATE $h(t) \gets W_0 x(t) + h^{\text{MoD}}(t)$
\ENDFOR

\STATE \textbf{// Load Balancing Loss for current layer (Training Only)}
\STATE $f^A_i = \frac{1}{T} \sum_{t=1}^{T} 1\left( i \in \mathcal{I}^A(t) \right)$
\STATE $P^A_i = \frac{1}{T} \sum_{t=1}^{T}\ell^A_i{(t)}$
\STATE $f^B_i = \frac{1}{T} \sum_{t=1}^{T} 1\left( i \in \mathcal{I}^B(t) \right)$
\STATE $P^B_i = \frac{1}{T} \sum_{t=1}^{T}\ell^B_i{(t)}$
\STATE $\mathcal{L}_{\text{aux\_layer}}$$ = \frac{E}{2} \sum_{i=1}^{E} \left( f^A_i \cdot P^A_i + f^B_i \cdot P^B_i \right)$
\STATE \RETURN $H$, $\mathcal{L}_{\text{aux\_layer}}$
\end{algorithmic}
\end{algorithm}

Algorithm ~\ref{alg:self_debate} details the concrete implementation of self-debate in MoD during inference. At the $K$-th reasoning round, the model revisits its own responses generated in the previous $K - 1$ rounds, which are explicitly incorporated into the prompt and concatenated with a dedicated self-debate instruction—specifically, “Review the following responses from other assistants and determine your final answer.” This formulation enables the model to effectively simulate a multi-agent debate process within a single model.

\begin{algorithm}
\caption{Iterative Self-Debate Pipeline}
\label{alg:self_debate}
\begin{algorithmic}[1]
\REQUIRE Image $I$, User Question $Q$
\REQUIRE Debate Prompt $\mathcal{D}$
\REQUIRE Multimodal LLM $\mathcal{M}$
\REQUIRE  Max Rounds $K$

\ENSURE Response Sequence $\mathcal{R} = \{R_0, R_1, \dots, R_K\}$

\STATE \textbf{// Phase 1: Generate Initial Response}
\STATE $C_{init} \gets \{I, Q\}$
\STATE $R_0 \gets \mathcal{M}(C_{init})$
\STATE Initialize history accumulator: $\mathcal{H} \gets R_0$
\STATE Add $R_0$ to result set $\mathcal{R}$

\STATE \textbf{// Phase 2: Iterative Self-Debate}
\FOR{$k = 1$ to $K$}
    \STATE \textbf{// Construct independent context with full history}
    \STATE $C_{k} \gets \{I, Q, \mathcal{D}, \mathcal{H}\}$
    \STATE \textbf{// Generate debate response}
    \STATE $R_k \gets \mathcal{M}(C_{k})$
    \STATE \textbf{// Accumulate history}
    \STATE $\mathcal{H} \gets \mathcal{H} \oplus R_k$
    \STATE Add $R_k$ to result set $\mathcal{R}$
\ENDFOR

\STATE \RETURN $\mathcal{R}$
\end{algorithmic}
\end{algorithm}

\section{Experimental Details}
\label{app:experimental}

\subsection{Implementation Details}
\label{app:implementation}

We build MoD upon LLaVA-v1.6-Vicuna-13B~\cite{liu2024llavanext} and Qwen2.5VL-3B-Instruct~\cite{wang2024qwen2}, two large vision language models with strong zero-shot reasoning capabilities.
The base model weights are kept frozen during fine-tuning, and only the dialectical expert modules and routing networks are optimized.
We implement our framework using PyTorch and conduct all experiments on NVIDIA A100 GPUs.
Table~\ref{tab:hyperparams-llava} and Table~\ref{tab:hyperparams-qwen} list the fine-tuning hyperparameters.

\begin{table}[t]
\centering
\small
\caption{Fine-tuning hyperparameters for llava}
\label{tab:hyperparams-llava}
\begin{tabular}{ll}
\toprule
\textbf{Hyperparameter} & \textbf{Value} \\
\midrule
Base Model & LLaVA-v1.6-Vicuna-13B \\
MoD target layer & $\{W_Q, W_K, W_V, W_O\}$ \\
Number of Experts $E$ & 8 \\
Activated Experts $r$ & 4 \\
Scaling Factor $\alpha$ & 8 \\
Sliding Window Size $\mathcal{W}$ & 16 \\
Learning Rate & $2 \times 10^{-5}$ \\
LR Schedule & Cosine Annealing \\
Batch Size & 8 \\
Training Epochs & 1 \\
Aux Loss Weight $\lambda$ & 0.01 \\
Optimizer & AdamW \\
Weight Decay & 0.01 \\
Warmup Ratio & 0.03 \\
Precision & FP16 \\
\bottomrule
\end{tabular}
\end{table}

\begin{table}[t]
\centering
\small
\caption{Fine-tuning hyperparameters for qwen}
\label{tab:hyperparams-qwen}
\begin{tabular}{ll}
\toprule
\textbf{Hyperparameter} & \textbf{Value} \\
\midrule
Base Model & Qwen2.5VL-3B-Instruct \\
MoD target layer & $\{W_Q, W_K, W_V, W_O\}$ \\
Number of Experts $E$ & 8 \\
Activated Experts $r$ & 4 \\
Scaling Factor $\alpha$ & 8 \\
Sliding Window Size $\mathcal{W}$ & 16 \\
Learning Rate & $2 \times 10^{-5}$ \\
LR Schedule & Cosine Annealing \\
Batch Size & 8 \\
Training Epochs & 1 \\
Aux Loss Weight $\lambda$ & 0.01 \\
Optimizer & AdamW \\
Weight Decay & 0.01 \\
Warmup Ratio & 0.03 \\
Precision & BF16 \\
\bottomrule
\end{tabular}
\end{table}

\begin{figure*}[t]
    \centering
    \includegraphics[width=\linewidth]{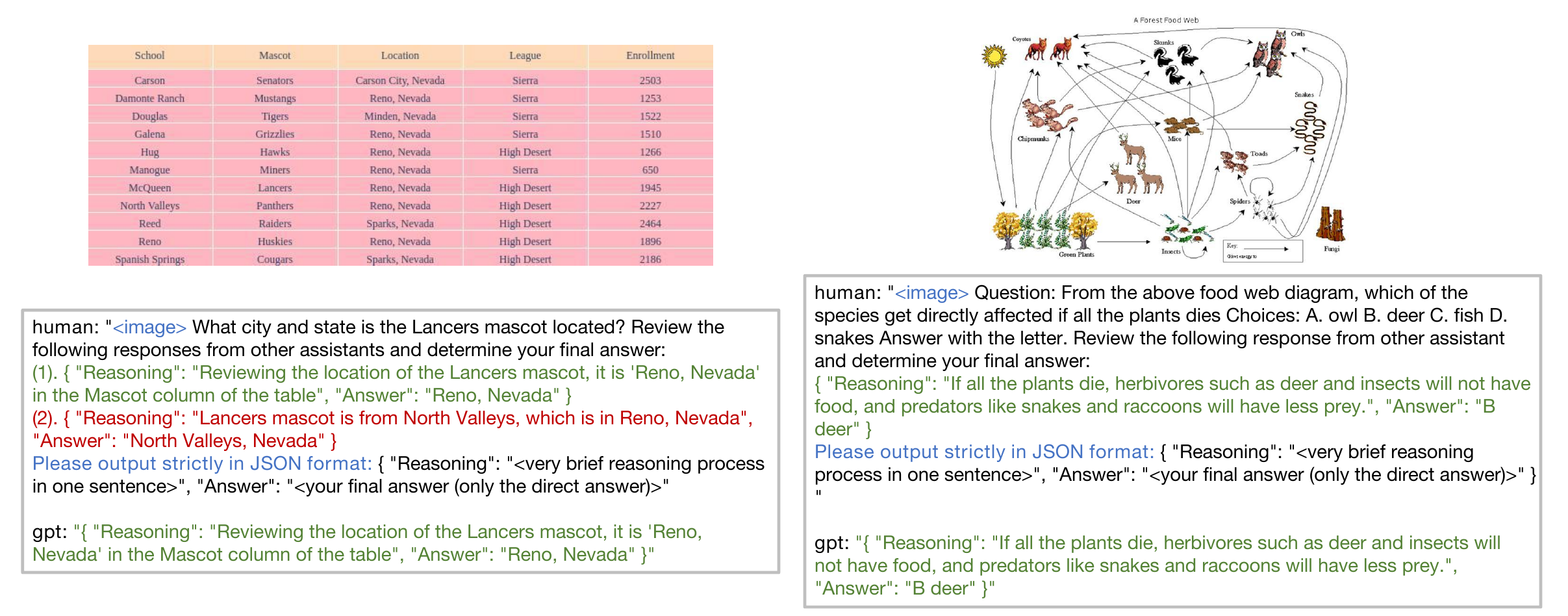}
    \Description{Examples of viewpoint-shift training data. Left: $\mathcal{T}_{rev}$ trajectory where the model reviews conflicting responses and corrects errors. Right: $\mathcal{T}_{pos}$ trajectory where the model confirms correct prior reasoning. Green highlights indicate correct reasoning; red highlights indicate errors to be revised.}
    \caption{Examples of viewpoint-shift training data. Left: $\mathcal{T}_{rev}$ trajectory where the model reviews conflicting responses and corrects errors. Right: $\mathcal{T}_{pos}$ trajectory where the model confirms correct prior reasoning. Green highlights indicate correct reasoning; red highlights indicate errors to be revised.}
    \label{fig:data_format}
\end{figure*}

\subsection{Tuning Data Format}
\label{app:data_format}

Figure~\ref{fig:data_format} illustrates our viewpoint-shift training data format. Each sample contains a visual input, a question, prior responses from previous debate rounds, and an instruction to produce a structured JSON output.

\paragraph{Correction and Revision ($\mathcal{T}_{rev}$).}
The left example shows a table-based reasoning task. The model receives two conflicting prior responses: (1) correctly identifies "Reno, Nevada" from the Mascot column; (2) incorrectly answers "North Valleys, Nevada" based on a different row. The model must evaluate both responses and produce the correct final answer by grounding in the visual evidence.

\paragraph{Consistent Reasoning ($\mathcal{T}_{pos}$).}
The right example shows a food web diagram. The prior response correctly reasons that if all plants die, herbivores like deer and insects will lack food, leading to "B deer" as the answer. The model confirms this reasoning and maintains the correct conclusion.

\paragraph{Format Structure.}
Each training sample follows the template: (1) \texttt{<image>} token referencing the visual input; (2) the original question; (3) prior assistant responses as context for review; (4) instruction to output in JSON format with \texttt{Reasoning} and \texttt{Answer} fields. This format trains the model to critically evaluate prior reasoning, identify errors when present, and produce grounded final answers.

\subsection{Benchmarks}
\label{app:datasets}

We evaluate on four benchmarks spanning diverse multimodal reasoning challenges:

\begin{itemize}
    \item \textbf{ScienceQA}~\cite{lu2022learn}: Multimodal science question answering across natural, social, and language sciences. We report accuracy on both VAL and TEST splits.
    
    \item \textbf{MMMU}~\cite{yue2024mmmu}: College-level reasoning involving spatial relationships, abstract concepts, and expert-level knowledge. DEV contains 150 samples and VAL contains 900 samples.

    \item \textbf{MMStar}~\cite{chen2024we}: Vision-indispensable multimodal benchmark designed to reduce language bias and test whether models truly use visual evidence. It covers diverse perception-and-reasoning scenarios; we report overall accuracy.
    
    \item \textbf{POPE}~\cite{lievaluating}: Evaluates object hallucination through binary yes/no questions about object existence in images.

    \item \textbf{MME}~\cite{fu2025mme}: Comprehensive evaluation covering perception abilities and cognition abilities. We report perception and cognition scores separately.

    \item \textbf{MMLU}~\cite{hendryckstest2021}: A text-only benchmark spanning 57 subjects across STEM, humanities, social sciences, and professional domains, used to verify that multimodal adaptation does not degrade language reasoning capabilities.
    
\end{itemize}

\subsection{Baselines}
\label{app:baselines}

We compare against three reasoning paradigms:

\begin{itemize}
    \item \textbf{Single Model Reasoning}: Standard zero-shot inference using LLaVA-v1.6-Vicuna-13B~\cite{liu2023llava} or Qwen2.5VL-3B-Instruct~\cite{bai2023qwen}.
    
    \item \textbf{Multi-Agent Debate (MAD)}~\cite{liang2024encouraging}: Instantiates three independent model instances (two debaters and one judge) for structured argumentation over multiple rounds.
    
    \item \textbf{Self-Correction (SC)}~\cite{he2025self}: A self-refinement approach where the model critiques and revises its initial response.
\end{itemize}

For LLaVA-v1.6, we evaluate all baselines and both MoD variants (MoD-Single and MoD-Debate).
For Qwen2.5VL-3B-Instruct, we apply MoD to demonstrate generalizability across different architectures and scales.
All comparisons within each base model use identical configurations for fair evaluation.

\subsection{Evaluation Protocol}
\label{app:evaluation}

We evaluate MoD in two inference modes:

\begin{itemize}
    \item \textbf{MoD-Single}: Single-round inference with MoD.
    
    \item \textbf{MoD-Multi}: Two-round self-debate within a single model instance, achieving effective self-correction with significantly lower token cost than MAD.
\end{itemize}

For multiple-choice questions (MMMU, ScienceQA), we compute exact match accuracy.
For MME, we report perception and cognition scores separately.
All results are averaged over three random seeds.

\section{Additional Expert Analysis}
\label{app:expert}

\subsection{Topic-wise Expert Activation}
\label{app:topic_expert}

\begin{figure*}
    \centering
    \includegraphics[width=\textwidth]{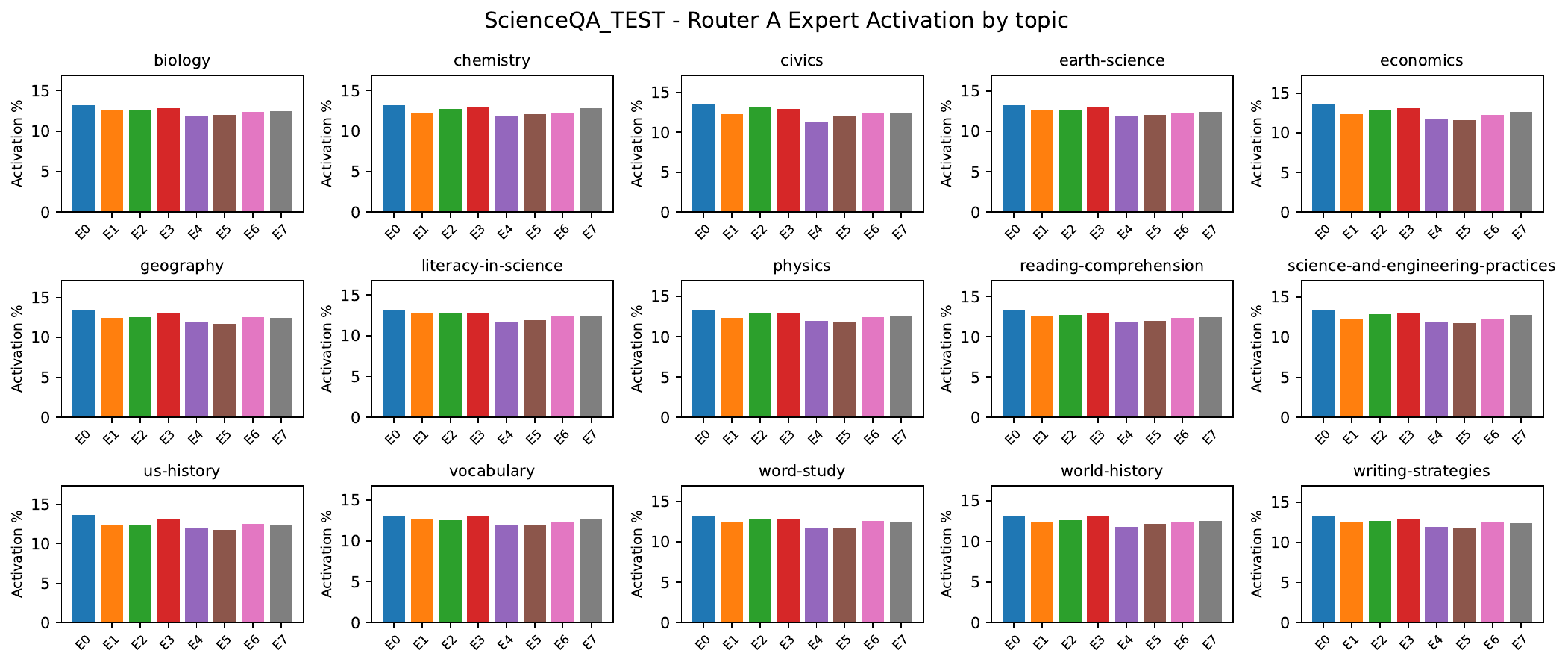}
    \includegraphics[width=\textwidth]{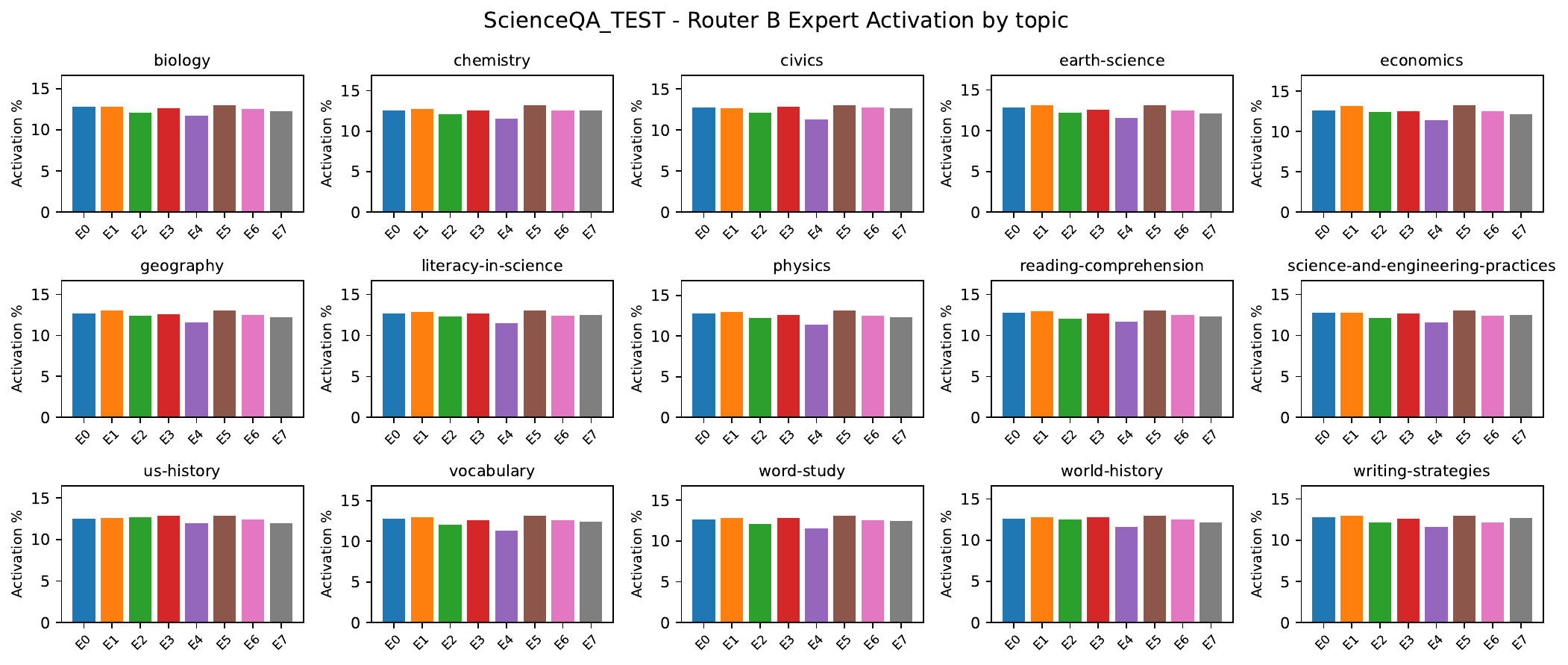}
    \Description{Router-A expert activation by topic on ScienceQA-TEST. Each subplot shows the activation percentage for all 8 experts within a specific topic. Activation remains balanced across experts while exhibiting subtle topic-dependent variations.}
    \caption{Router-A expert activation by topic on ScienceQA-TEST. Each subplot shows the activation percentage for all 8 experts within a specific topic. Activation remains balanced across experts while exhibiting subtle topic-dependent variations.}
    \label{fig:topic_expert_detail}
\end{figure*}

Figure~\ref{fig:topic_expert_detail} presents detailed expert activation distributions for Router-A across all 15 topics in ScienceQA-TEST.
While overall activation remains balanced due to our load-balancing objective, subtle topic-dependent patterns emerge.
For example, E0 shows consistently high activation across most topics, while E4 exhibits relatively lower activation.
Science-related topics (biology, chemistry, physics) show similar activation profiles, as do humanities topics (us-history, world-history), suggesting that the router learns to group semantically related categories.

\subsection{Router Comparison Heatmap}
\label{app:router_heatmap}

\begin{figure*}
    \centering
    \includegraphics[width=0.85\textwidth]{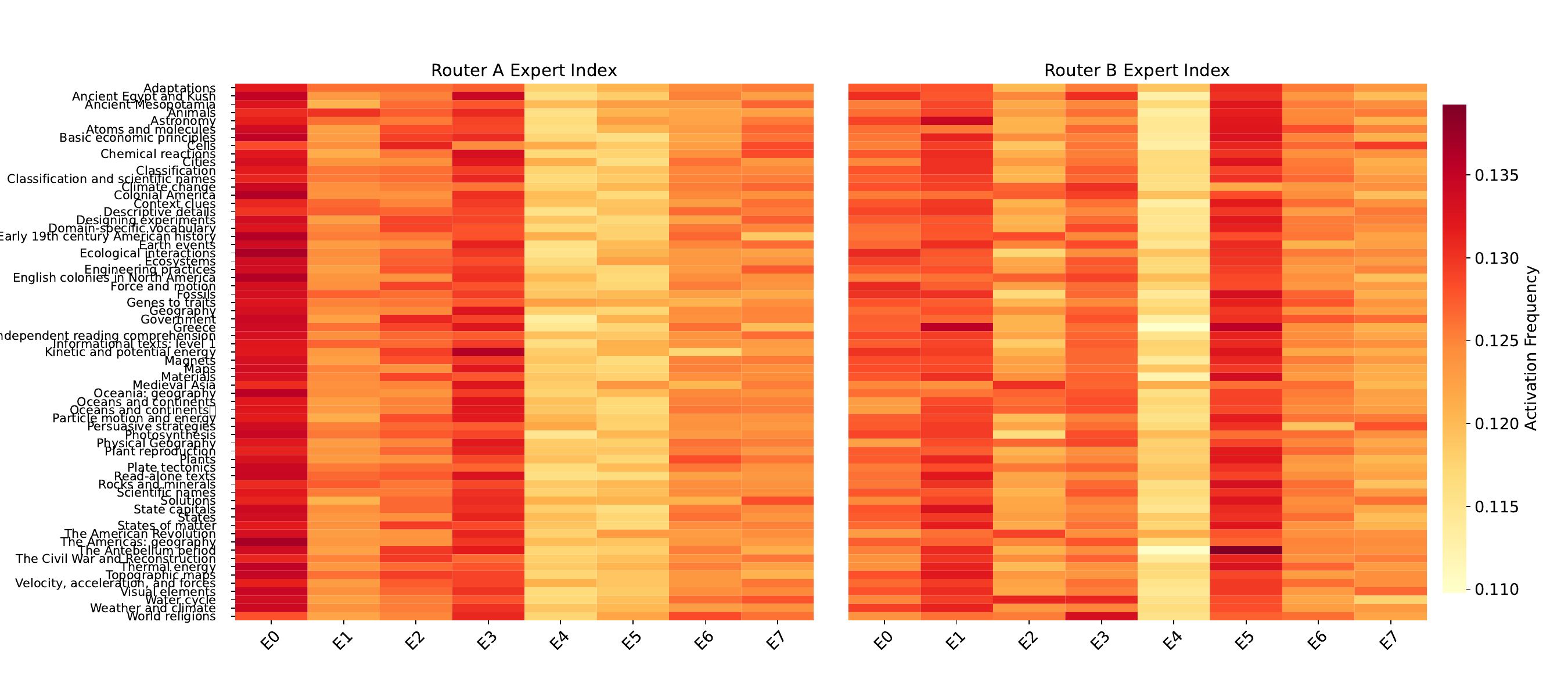}
    \Description{Expert activation frequency comparison between Router-A (A-side) and Router-B (B-side) on ScienceQA-TEST. Router-A exhibits higher concentration on specific experts, while Router-B shows more uniform distribution.}
    \caption{Expert activation frequency comparison between Router-A (A-side) and Router-B (B-side) on ScienceQA-TEST. Router-A exhibits higher concentration on specific experts, while Router-B shows more uniform distribution.}
    \label{fig:router_heatmap}
\end{figure*}

Figure~\ref{fig:router_heatmap} provides a direct comparison of expert activation frequencies between Router-A (role allocation) and Router-B (process control) on ScienceQA-TEST.
Router-A shows higher activation concentration on E0 (approximately 13\%), while Router-B distributes more uniformly across experts (all within 12.2\%-12.7\%).
This divergence validates that our dual-routing mechanism learns differentiated strategies: Router-A develops stronger expert preferences for role-specific processing, while Router-B maintains broader coverage for flexible synthesis.

\subsection{Router Activation Patterns}
\label{app:router_patterns}

\begin{figure*}
    \centering
    \includegraphics[width=\linewidth]{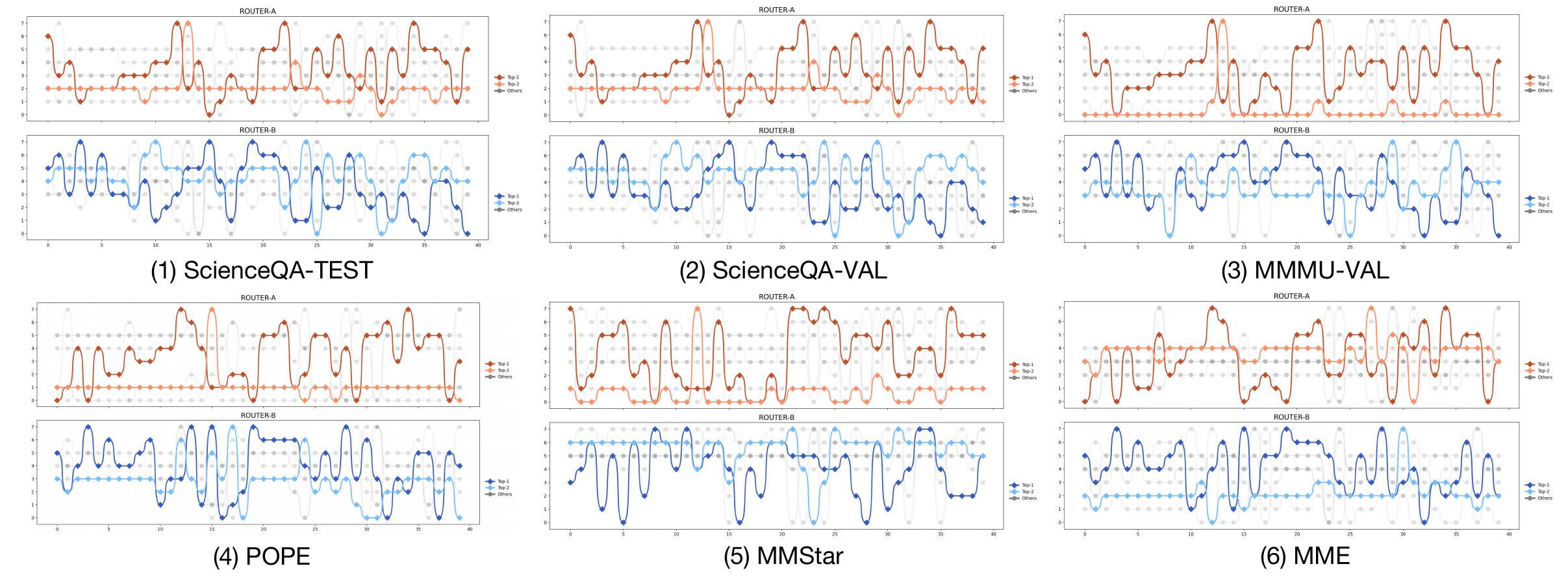}
    \Description{Layer-wise router activations across benchmarks. Orange: Router-A; Blue: Router-B. Smooth curves validate momentum switching; divergent A/B patterns confirm learned specialization.}
    \caption{Layer-wise router activations across benchmarks. Orange: Router-A; Blue: Router-B. Smooth curves validate momentum switching; divergent A/B patterns confirm learned specialization.}
    \label{fig:router_patterns}
\end{figure*}

Figure~\ref{fig:router_patterns} visualizes layer-wise routing activations for Router-A (orange) and Router-B (blue) across six benchmarks. Both routers exhibit smooth, continuous activation patterns rather than erratic switching, validating that opinion-level switching prevents token-level fluctuations. Router-A maintains relatively stable patterns across benchmarks, while Router-B shows more pronounced variations in deeper layers, adapting synthesis strategies to task requirements. ScienceQA-TEST and ScienceQA-VAL show highly similar patterns within the same task domain. MMMU-VAL exhibits sharper Router-B transitions for complex reasoning. POPE displays smoother patterns consistent with binary classification. MMStar and MME show more dynamic Router-B activations for diverse question types.

\subsection{Effect of auxiliary Loss}
\label{app:load_balance}

\begin{figure*}
    \centering
    \includegraphics[width=\linewidth]{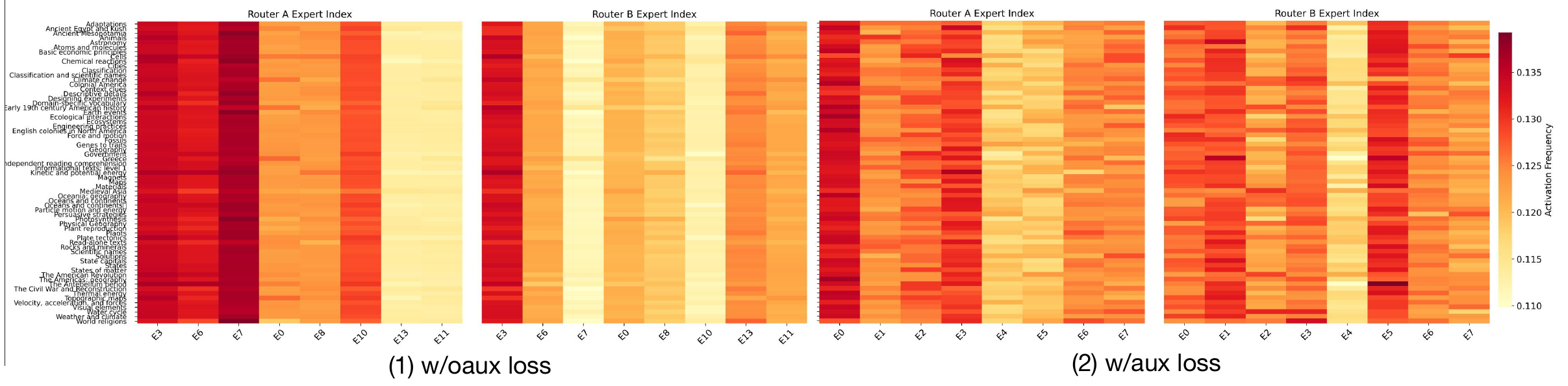}
    \Description{Expert activation heatmaps without (left) and with (right) auxiliary loss in ScienceQA-TEST. Without the auxiliary loss, activation concentrates on few experts; with the auxiliary loss, utilization is balanced across all experts.}
    \caption{Expert activation heatmaps without (left) and with (right) auxiliary loss in ScienceQA-TEST. Without the auxiliary loss, activation concentrates on few experts; with the auxiliary loss, utilization is balanced across all experts.}
    \label{fig:load_balance}
\end{figure*}

Figure~\ref{fig:load_balance} compares expert activation patterns with and without the auxiliary loss. Without the loss (left), expert selection collapses to a narrow subset, with E2 and E5 dominating across all topics while other experts remain underutilized. With the loss (right), activation distributes more evenly across all experts, preserving the full $N \times N$ combinatorial capacity. This validates that the auxiliary loss is essential for preventing expert collapse and maintaining diverse reasoning pathways.

\end{document}